# Large-scale semantic mapping of learner agency and autonomy reveals what measurement and generative AI research overlook

Fei Qin[1#], Xiaobo Liu[1#], Yaowen Zhang[1], Xuming Li[1], Fei Wang[2], Mutlu Cukurova[3], Jingjing Chen[1*], Yu Zhang[1*]

[1] School of Education, Tsinghua University, Beijing 100084, China

[2] Department of Psychological and Cognitive Sciences, Tsinghua University, Beijing 100084, China

[3] Institute of Education, University College London, London, UK

(# These authors contributed equally to this work.)

Corresponding authors:

Yu Zhang, zhangyu2011@tsinghua.edu.cn

Jingjing Chen, jingjing-chen@tsinghua.edu.cn

## Abstract

Learner agency and autonomy are foundational to personal development, yet a pervasive "jingle-jangle" fallacy (i.e. identical terms denoting different constructs, distinct terms denoting identical ones) has substantially hindered cumulative knowledge. Treating meaning as a phenomenon constituted through use in linguistic practice, we extracted 8,954 definitions and 2,700 scale items from over 14,000 publications, to investigate how researchers actually used learner agency and autonomy with a semantic analysis pipeline. The definitional landscape of two constructs resolves into three dimensions: regulation and control of learning (task), intrinsic motivation and internal decision-making (person), and social-relational action (sociocultural), thereby empirically quantifying the jingle-jangle fallacy. Existing scales, however, systematically underrepresent the sociocultural dimension. Critically, current generative AI research in education concentrates on learning regulation and control, narrowing the behavioral repertoire that AI-mediated learning environments are designed to cultivate. Beyond conceptual clarification, this work carries direct implications for conceptualization, measurement, and practice towards supporting the multidimensional learner agency and autonomy.

## Introduction

In the era of generative artificial intelligence (GenAI), human agency has become a renewed concern, as GenAI systems increasingly participate in cognitive work, decision support, and persuasive communication, thereby reshaping the conditions under which individuals form judgments and exercise autonomous choice (Bai et al., 2025; Krakowski, 2025; Matz et al., 2024; Salvi et al., 2025). International organizations have increasingly emphasized the critical importance of empowering students to regulate their own learning, adapt to complex social environments, and actively steer their future trajectories alongside AI (OECD, 2019, 2025; UNESCO, 2021, 2023, 2024). Within this broader societal context, learner agency and learner autonomy represent foundational capacities for sustaining human judgment, adaptive participation, and meaningful self-direction in an AI-mediated world. Over decades, these constructs have attracted sustained academic attention across multiple disciplines, including education, psychology, and sociology (Bandura, 1989, 2001; Benson, 2007; Biesta & Tedder, 2007; Cukurova, 2026; Deci & Ryan, 1987; Holec, 1981; Little, 1995; Moore, 1972; Reeve & Tseng, 2011; Teng, 2019; Viberg et al., 2026). This long-standing cross-disciplinary inquiry has generated over hundreds of thousands of studies, offering a rich theoretical and empirical foundation for examining these constructs.

Despite this apparent abundance, learner agency and learner autonomy remain conceptually confused. This phenomenon is commonly described as the "jingle-jangle fallacy" (Kelley, 1927; Van Petegem et al., 2013). In particular, a "jingle" problem arises when the same construct label is used to refer to different underlying constructs. For example, the construct label in both of the following cases is "learner agency." One focuses on agency as relationally constituted through social engagement over time: "temporally embedded process of social engagement" (Emirbayer & Mische, 1998, p. 963), stressing its relational emergence through social interaction across time. The other, by contrast, treats it as an internally generated primal driving force: "the power to originate action for given purposes" (Bandura, 1997, p. 3; see also Hang et al., 2025, p. 20), stressing purely psychological endogenous intention and executory capacity. A "jangle" problem, by contrast, arises when different construct labels are used to denote highly similar or overlapping constructs. For example, "learner agency" has been defined as "the capacity to act independently and to make one's own choices" (Manyukhina & Wyse, 2019, p. 223). At the same time, "learner autonomy" has been defined in a highly similar manner as "a capacity for thinking and acting independently that may occur in any kind of situation" (Littlewood, 1996, p. 428; see also Lin & Reigeluth, 2019, p. 2670).

Such confusion is not a trivial semantic inconvenience. The jingle-jangle fallacies would undermine construct validity, weaken discriminant validity, and compromise the comparability of empirical findings (Cronbach & Meehl, 1955; Gonzalez et al., 2021). In the long run, these ambiguities deepen epistemological fragmentation across disciplines, obstruct interdisciplinary dialogue, and slow the cumulative development of coherent scientific knowledge (Marsh, 1994). Moreover, as generative AI reshapes the conditions of human development worldwide, learner agency and autonomy have moved from theoretical constructs to active design imperatives (Kim et al., 2025). Decisions about how to support learner agency and autonomy hinge on assumptions about what these constructs mean. Without clear definitions of learner agency and autonomy, there is no way to evaluate whether any given intervention is supporting them, ignoring them, or actively undermining them.

Traditionally, resolving such conceptual entanglements has mainly relied on normative, top-down analysis, which attempts to prescribe definitions based on theoretical lineage (Code, 2020; Gerring, 1999; Goertz, 2006). However, these theoretical efforts often fail to converge because the discourse is trapped in "disciplinary silos" (Jacobs, 2014), where definitions remain self-referential and rooted in incommensurable operationalizations. Furthermore, when the literature reaches a scale of hundreds of thousands of publications, the jingle-jangle fallacy is no longer a mere philosophical disagreement that can be settled by better logic. It becomes a structural property of the field's linguistic reality. Purely theoretical distinctions thus become disconnected from how these constructs are actually used in practice. To break this deadlock, we need to move beyond asking how these constructs should differ in principle and instead investigate how they do behave within the massive, collective semantic space they inhabit. To achieve this, we ground our methodology in functionalist linguistics and distributional semantics. This approach posits that meaning is not an inherent, static property of a word, but rather a dynamic phenomenon constituted through continuous, patterned use in linguistic practice (Boleda, 2020; Fedorenko et al., 2024; Harris, 1954; Wittgenstein, 1953). From a usage-based perspective, a construct's practical definition is revealed by its linguistic environment, including the specific syntactic structures, collocations, and co-occurrences it consistently attracts. When researchers write about constructs, they leave behind empirical, statistical traces of their conceptualizations. By examining the distributional patterns and relational structures of "learner agency" and "learner autonomy" within this large-scale definitional landscape, we can identify the implicit, collective consensus of the field. Ultimately, this allows us to shift the lens from rigid, prescriptive debates toward a bottom-up, data-driven perspective, observing the natural boundaries and overlaps of these constructs as they are actually negotiated in academic discourse.

The operationalization of this bottom-up perspective is made possible by recent breakthroughs in Large Language Model (LLM)-based semantic embedding. Unlike traditional content analysis, which is limited by human labor resources and sampling bias, semantic embedding allows us to analyze the "population" of academic definitions rather than mere fragments (Hussain et al., 2024; Wulff & Mata, 2025a; Wulff & Mata, 2025b). By transforming complex definitions into high-dimensional vectors, the similarity between constructs can be precisely quantified (e.g., with cosine similarity), recasting abstract conceptual debates into a measurable geometric problem in this shared semantic space (Hussain et al., 2024; Wulff & Mata, 2025a). Recent pioneering studies have successfully applied this technique to reveal taxonomic incommensurability in personality psychology measurement (Wulff & Mata, 2025b), analyze the representation of complex constructs in internet discourse (Dorison & Charlesworth, 2025), and reduce redundant measurement in mental disorder scales (Voppel et al., 2025).

Building on methodological evidence that semantic embeddings can detect taxonomic incommensurability (Wulff & Mata, 2025b), the present study aimed to characterize learner agency and learner autonomy: two constructs that have resisted theoretical resolution across multiple disciplines for decades, and whose conceptual clarity is now critical for guiding GenAI-driven educational design. Given that research on these constructs has produced large-scale, diffuse literatures in which comprehensive human review is infeasible, we designed an LLM-supported pipeline that automatically retrieves tens of thousands of relevant publications, extracts how the constructs are defined and measured, and synthesizes variations across studies based on semantic embeddings. The pipeline operates entirely through commercial LLM APIs, requiring no custom model training, and is thus accessible to scholars without deep machine learning expertise. By projecting fragmented theories onto a shared semantic space, the present study provides an objective benchmark for identifying redundancies and disentangling the conceptual landscapes of learner agency and autonomy. Specifically, it first maps where, to what degree, and along which dimensions the two constructs converge and diverge within a shared semantic space; then, based on this definitional analysis, the study further examines whether existing scales adequately cover the semantic content of the constructs they intend to measure and whether measures of the two constructs remain semantically distinguishable from one another; finally, the study applies this framework to GenAI-related research, demonstrating how such clarification can reveal hidden imbalances in emerging technological fields. By doing this, this study aims to replace decades of unresolved theoretical debate with measurable coordinates, enabling better interdisciplinary communication, more precise operationalization, and more intentional educational practices that account for the complexity of learner agency and autonomy.

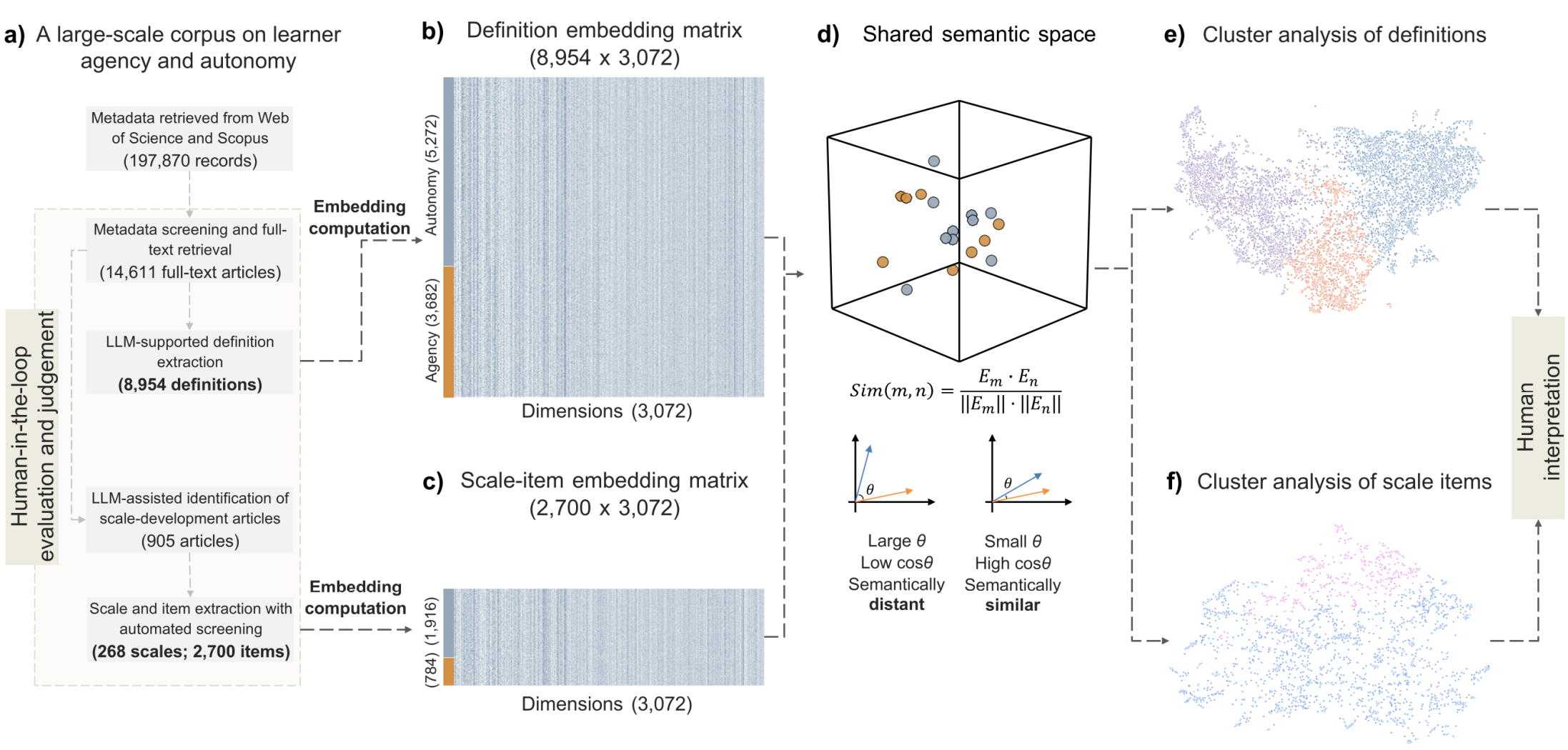

**Figure 1. Overview of the automated construct synthesis approach using semantic embeddings. (a)** Workflow for constructing a large-scale corpus on learner agency and autonomy. **(b-c)** Embedding matrices for definitions and scale items. Each row represents a construct definition or a scale item, and each column corresponds to one dimension of the embedding vector. **(d)** Shared semantic space constructed from the embeddings. After embedding computation, both definitions and scale items can be analyzed within the high-dimensional semantic space. For any two text segments, semantic similarity is quantified using cosine similarity between their embedding vectors. A smaller angle $\theta$ and larger cos$\theta$ indicate higher semantic similarity, whereas a larger angle $\theta$ and smaller cos$\theta$ indicate greater semantic distance. **(e-f)** Cluster analyses of definitions and scale items based on semantic embeddings. By leveraging mathematical similarity within the shared semantic space, clustering analysis identifies latent semantic structures, which are subsequently interpreted through human expert judgment.

# Results

Figure 1 presents an overview of the analytic pipeline. To systematically quantify the conceptual confusion between agency and autonomy within the semantic landscape, we first assembled a large-scale full-text corpus. This corpus encompassed 14,611 full-text academic articles discussing learner agency and autonomy from major literature databases (see Supplementary Information section *Literature Corpus Construction* for details). From this corpus, we extracted 8,954 English-language definitions using a long-context LLM pipeline that identified explicit learner- or student-centered definitions while preserving the original source wording, yielding 5,272 definitions classified as autonomy and 3,682 classified as agency. Extraction fidelity was verified through exact string matching with the source documents, showing that 95.5% of the definitions were reproduced verbatim (see Supplementary Information for full extraction validation). These definitions spanned publications from 1970 to 2026 and involved over 3,000 research institutions across 109 countries and regions, ensuring broad temporal and geographic coverage. For the extracted definitions, we utilized the *text-embedding-3-large* (OpenAI) model to generate semantic embeddings, projecting each definition into a 3,072-dimensional semantic space.

**Semantic Analysis Reveals Three Latent Dimensions Rather Than a Binary Agency-Autonomy Distinction**

Based on these high-dimensional vectors, we examined whether the definitions of learner agency and autonomy exhibited clear conceptual boundaries in the semantic landscape, or instead followed a different latent structure. We applied a data-driven clustering approach to explore the underlying semantic structure of the entire definitional corpus. We evaluated both agglomerative hierarchical clustering and K-means across multiple combinations of UMAP dimensionality-reduction hyperparameters and clustering configurations, selecting the optimal solution based on silhouette scores computed in the original embedding space (see Supplementary Information section *Clustering Analysis of Definitions* for details of the hyperparameter search). The best-performing solution was a three-cluster hierarchical partition (C1, C2, and C3), with 143 (1.6% of 8,954) definitions identified as outliers and excluded from subsequent analyses (see the details of outliers in Supplementary Information section).

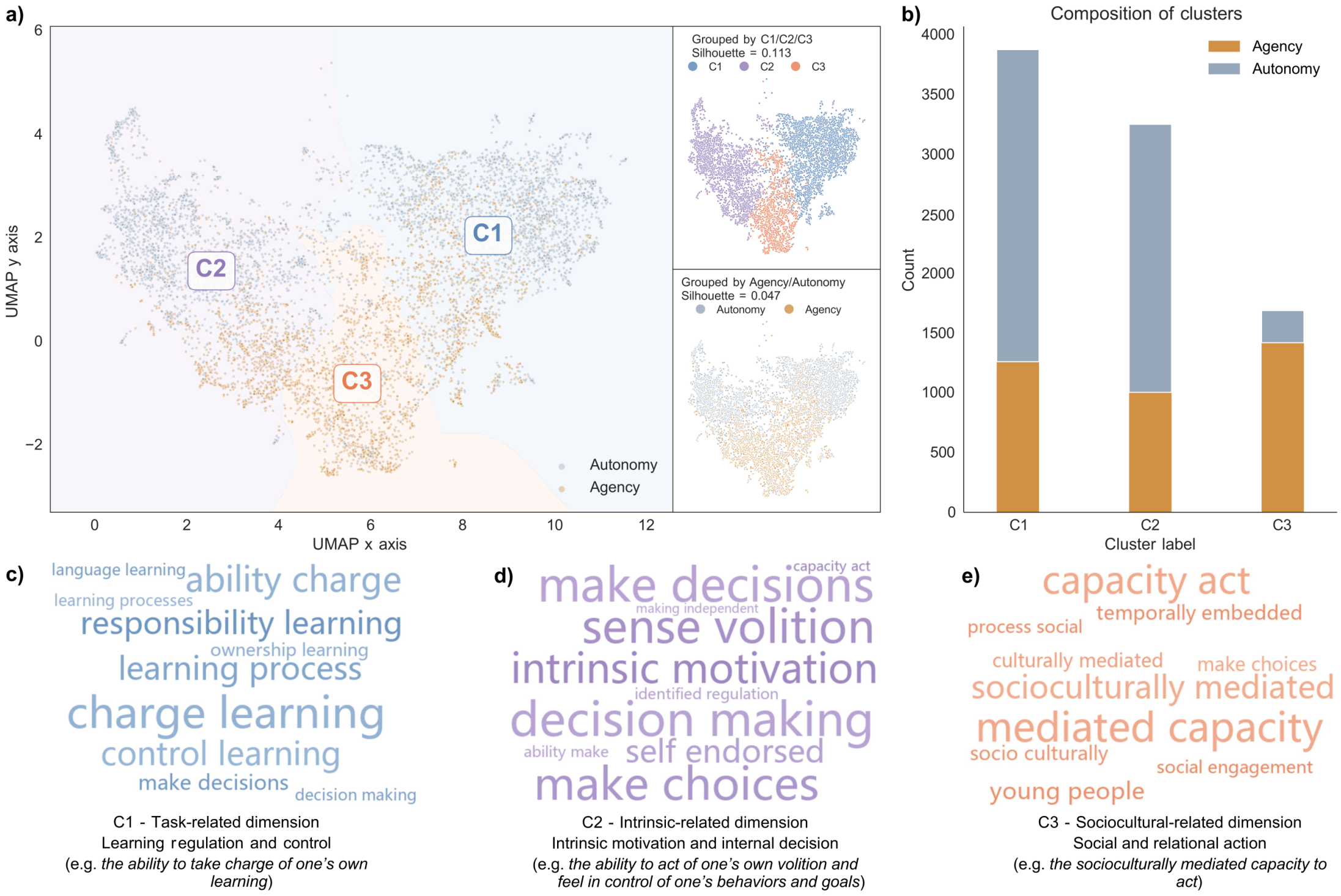

**Figure 2. Semantic structure of learner agency and autonomy definitions. (a)** UMAP projection of the semantic embeddings of the extracted definitions. In the main panel, the semantic regions associated with the three clusters were color-shaded by assigning the surrounding UMAP space to C1, C2, or C3 using a k-nearest neighbors classifier based on the clustering results. These three

clusters correspond to C1 (learning regulation and control, a task-related dimension), C2 (intrinsic motivation and internal decision, an intrinsic-related dimension), and C3 (social and relational action, a sociocultural-related dimension). Individual points are colored by their original construct labels, autonomy or agency. The two inset panels on the right display the same two-dimensional UMAP coordinates, grouped by cluster membership (C1/C2/C3) and by the original autonomy-agency labels, with silhouette scores annotated for each grouping. **(b)** Stacked bar chart showing the composition of each cluster by construct label (agency vs. autonomy). **(c-e)** Word clouds showing the most prominent terms in C1, C2, and C3, respectively, based on c-TF-IDF scores. Common English function words were removed before term-frequency computation, and all displayed terms are 2-grams.

As shown in the UMAP projection in Figure 2a, these three clusters occupied coherent and largely separable regions of the semantic landscape, in sharp contrast to the heavily intermixed distribution of the original autonomy and agency labels within the same space. This contrast was quantified by silhouette scores. The three-cluster solution (C1/C2/C3) yielded a substantially higher silhouette score (0.113) than the original binary autonomy-agency grouping (0.047). Although such absolute values are modest, they are well within the typical range reported for high-dimensional short-text embedding clustering, where continuously overlapping semantic boundaries tend to depress silhouette values (e.g., Muthusami et al., 2024; Abdalgader et al., 2024). The relative gap therefore indicates that the three-cluster partition captures the latent structure of the data markedly better than the conventional binary distinction.

We further compared the pairwise cosine similarities between embeddings under both groupings. Welch's t-tests, with $p$ values obtained from 5,000-iteration permutation tests that shuffled the embedding labels to account for non-independence among similarity pairs (hereafter the same), revealed a marked divergence between the two schemes. Under the binary labels, within-construct similarities were statistically higher than cross-construct similarities. The means for autonomy-autonomy pairs ($M$ = 0.399, $SD$ = 0.129) and agency-agency pairs ($M$ = 0.360, $SD$ = 0.107) both exceeded those for the autonomy-agency cross-construct pairs ($M$ = 0.349, $SD$ = 0.109), yet the largest of these differences was negligible in magnitude, with the agency-agency versus cross-construct contrast yielding a trivial effect size (Cohen's $d$ = 0.094) despite a permutation $p$ value below .001. Under the three-cluster scheme, by contrast, the same analysis recovered a far stronger separation, in which within-cluster similarities consistently and substantially exceeded between-cluster similarities, with effect sizes for these contrasts ranging from medium to large (Cohen's $d$ = 0.41–1.22, all permutation $p < .001$).

Taken together, these results demonstrate that the definitional landscape is not organized along a simple binary division between agency and autonomy, but rather structured across three coherent, latent dimensions. This distribution suggests that learner agency and autonomy function as broad, umbrella-like constructs covering a shared conceptual space. Rather than marking two discrete theoretical territories, these labels each span overlapping portions of the same three latent dimensions. This directly evidences the jingle fallacy at work, whereby the same label, whether "learner agency" or "learner autonomy", is used to refer to conceptually distinct component, further reinforcing the conclusion that the conventional binary distinction fails to capture the deeper semantic organization of the field.

We now turn to the specific composition of each of the three clusters. Cluster C1 ($n$ = 3,873;

agency = 1,255 [32.4%], autonomy = 2,618 [67.6%]) represents the task-related latent dimension of "learning regulation and control." As revealed by the high-frequency semantic features extracted using the c-TF-IDF algorithm, the salience of this dimension is primarily marked by core expressions such as "[to take] charge [of one's] learning," "control [of] learning," "ability [to take] charge," and "responsibility [for] learning" (Figure 2c). These terms reveal a task-specific semantic focus, indicating that the texts emphasize the management, guidance, and responsibility for the learning activity itself. To identify the most representative definitions within each cluster, we computed the cosine similarity between each definition embedding and the cluster centroid (the average embedding vector within the cluster) and ranked definitions accordingly (Supplementary Table S1). The definitions closest to the C1 centroid all contained similar expressions centering on "the ability to take charge of one's own learning." This formulation was also the most frequently repeated definition in the corpus (originally from Holec, 1981, p. 2, appearing 61 times) and can thus be regarded as the most representative definition of this dimension.

Cluster C2 (*n* = 3,250; agency = 1,000 [30.8%], autonomy = 2,250 [69.2%]) reflects the intrinsic-related latent dimension of "intrinsic motivation and internal decision." Its core marker vocabulary includes "make decisions," "sense [of] volition," "intrinsic motivation," and "make choices" (Figure 2d). Compared to the task-related nature of C1, this intrinsic-related cluster pivots toward a deeper focus on learners' internal motivation. Semantic analysis indicates that C2 strongly emphasizes self-endorsed action, the generation of internal decision, and individuals' perceived control over their own behaviors and goals. The most representative definition in C2 describes the focal construct as "the ability to act of one's own volition and feel in control of one's behaviors and goals," and the most frequently repeated definition was "a capacity for detachment, critical reflection, decision-making, and independent action" (originally from Little, 1991, p. 4, appearing 8 times).

In contrast, Cluster C3 (*n* = 1,688; agency = 1,419 [84.1%], autonomy = 269 [15.9%]) captures a distinct sociocultural-related latent dimension of "social and relational action." Its most salient semantic units comprise "capacity [to] act," "mediated capacity," "socioculturally mediated," "temporally embedded," and "social engagement" (Figure 2e). These terms signal a shift in theoretical perspective in which agency is constructed as a product dynamically emerging through specific contexts, tool mediation, and social relational networks. The most frequently repeated definition within this sociocultural-related dimension conceptualizes agency as a "socioculturally mediated capacity to act" (originally from Ahearn, 2001, p. 112, appearing 19 times), portraying it as a temporally embedded relational process synergistically shaped by the intertwining of past experiences, present conditions, and future aspirations.

**Revealing the Jangle Fallacy: The Unequal Distribution of Learner Agency and Autonomy Across Three Semantic Dimensions**

After identifying the three-cluster structure, we next sought to examine the relationships between the three semantic dimensions and the original labels of learner agency and learner autonomy. Specifically, we asked whether their distributions are comparable, and if not, which construct dominates the core semantics of each cluster. Two-dimensional UMAP visualizations of the definition embeddings within each of the three clusters were first generated, as shown in Figure 3a-c. Visually, the spatial positions occupied by autonomy and agency definitions did not fully overlap, suggesting that agency and autonomy retain distinct semantic structures within each cluster rather than being unified into a single undifferentiated space. To further analyze this relationship from a quantitative perspective, we compared the cosine similarity between each individual definition and the corresponding cluster centroid. As shown in Figure 3d-f, autonomy definitions were semantically closer to the centroid than agency definitions in both C1 (learning regulation and control; mean cosine similarity to centroid: agency − autonomy = −0.058, Welch's $t$ (2356.16) = −15.33, $p < .001$) and C2 (intrinsic motivation and internal decision; agency − autonomy = −0.034, Welch's $t$ (1920.89) = −8.73, $p < .001$). By contrast, C3 (social and relational action) showed the opposite orientation, with agency definitions being closer to the centroid than autonomy definitions (agency − autonomy = 0.061, Welch's $t$ (377.66) = 10.47, $p < .001$). This pattern remained robust when centroids were constructed separately from agency-only and autonomy-only definitions (Supplementary Figure S3). Together, these results indicate a structured asymmetry in how the two constructs contribute to each semantic dimension, with autonomy dominating C1 and C2 while agency dominates C3.

To further characterize the jangle fallacy of the two constructs, we computed pairwise cosine similarities between autonomy-autonomy, agency-agency, and autonomy-agency definition pairs within each of the three clusters separately (Figure 4g-i). These empirical results provide a critical quantitative lens through which to examine the jangle fallacy. Regarding Cluster C1, autonomy-autonomy pairs showed higher similarity than autonomy-agency pairs (mean: 0.489 vs. 0.433, Cohen's $d = 0.452$, permutation $p < .001$), whereas agency-agency pairs were indistinguishable from autonomy-agency pairs (mean: 0.428 vs. 0.433, Cohen's $d = −0.041$, permutation $p = .069$). This pattern indicates that the semantic core of C1 is organized almost entirely around autonomy, with agency definitions merely orienting toward this autonomy-centered structure rather than forming a coherent sub-cluster of their own. Consequently, within this task-related dimension (C1), the use of the label "agency" frequently constitutes a jangle fallacy by acting largely as a redundant descriptor for the underlying core of autonomy. Regarding Cluster C2, a more intermixed configuration emerged. Autonomy-autonomy pairs were more semantically similar than autonomy-agency pairs (mean: 0.435 vs. 0.395, Cohen's $d = 0.348$, permutation $p < .001$), and agency-agency pairs were also more semantically similar than autonomy-agency pairs (mean: 0.414 vs. 0.395, Cohen's $d = 0.175$, permutation $p < .001$), while autonomy-autonomy pairs were slightly more similar than agency-agency pairs (mean: 0.435 vs. 0.414, Cohen's $d = 0.172$, permutation $p < .001$). This suggests that both autonomy and agency definitions form relatively stable internal semantic structures within C2. Because a more nuanced boundary is maintained in C2, both autonomy and agency avoid complete semantic conflation, thereby resisting the jangle fallacy to a certain degree. Regarding Cluster C3, the pattern mirrored that of C1 but in the opposite direction. Agency-agency pairs were substantially more similar than autonomy-agency pairs (mean: 0.373 vs. 0.331, Cohen's $d = 0.426$, permutation $p < .001$), whereas autonomy-autonomy pairs were even slightly less similar than autonomy-agency pairs (mean: 0.324 vs. 0.331, Cohen's $d = −0.070$, permutation $p = .074$). Additionally, agency-agency pairs were substantially more similar than autonomy-autonomy pairs (mean: 0.373 vs. 0.324, Cohen's $d = 0.478$, permutation $p < .001$). This indicates that C3 is organized primarily around agency, with autonomy definitions lacking coherent internal convergence. Therefore, within this sociocultural-related dimension (C3), agency definitions dictate the semantic core. Labeling a definition as "autonomy" in this sociocultural context often masks its true semantic anchor, which is fundamentally agency, thus manifesting another clear instance of the jangle fallacy.

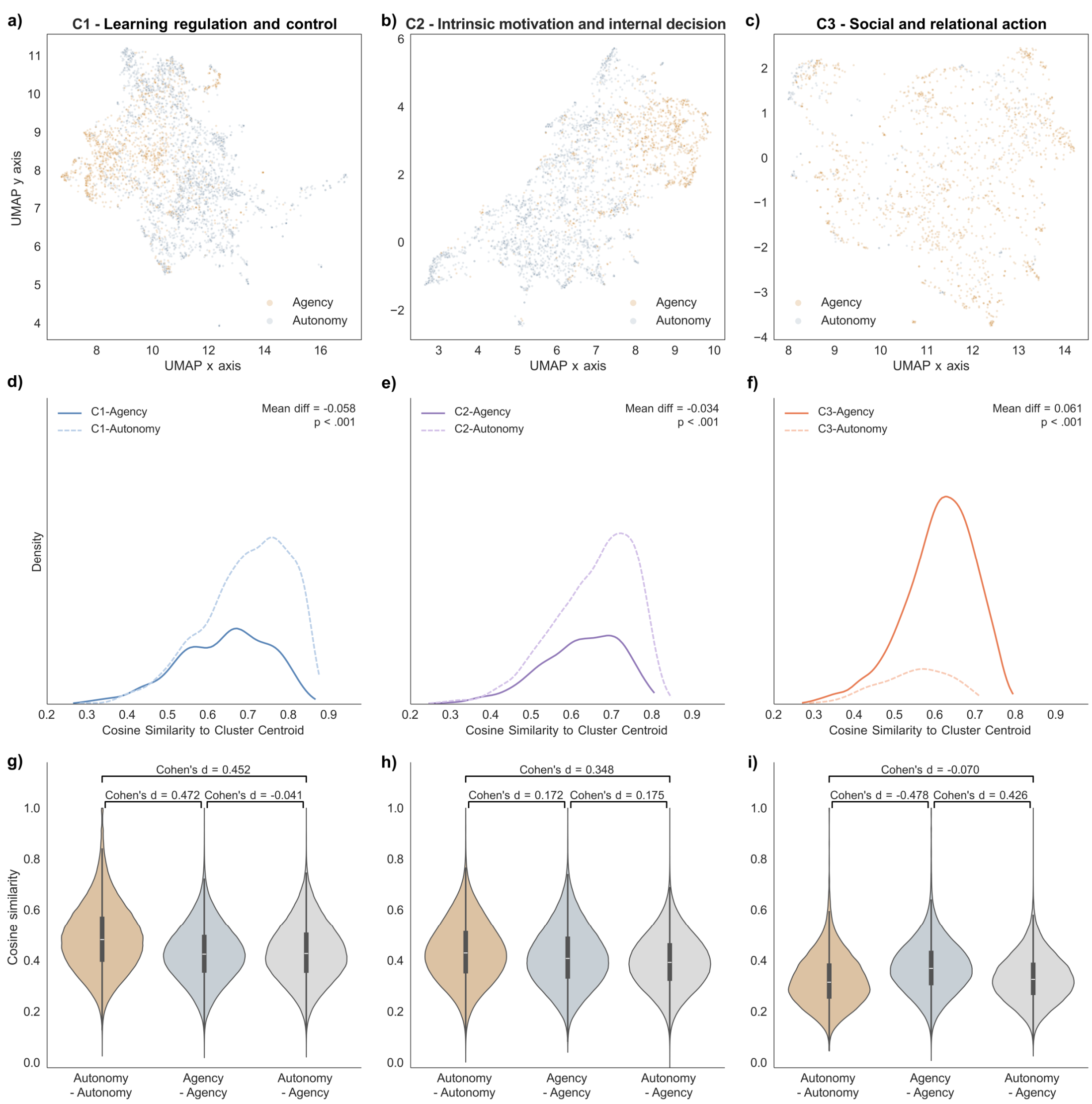

**Figure 3. Construct labels map asymmetrically onto latent semantic dimensions. (a-c)** UMAP projections of the semantic embeddings within the three clusters, shown separately for C1, C2, and C3. In each panel, agency and autonomy definitions are represented by points in different colors. **(d-f)** Within each cluster, density distributions of cosine similarity between individual definitions and the corresponding cluster centroid, plotted separately for agency and autonomy definitions. Mean differences between the two distributions and their significance levels are reported in each panel. **(g-i)** Violin plots showing the distribution of cosine similarity for three types of definition pairs within each cluster, including autonomy-autonomy, agency-agency, and autonomy-agency pairs. Cohen's d values were calculated to compare effect sizes among the three pair types within each cluster.

### Scales Reproduce Jingle-Jangle Patterns and Show Limited Content Validity

We then applied the same analytical approach to scale-based measurement to examine whether the jingle-jangle fallacy manifests at the level of operationalization. Specifically, we extracted 2,700 scale items from 268 instruments designed to measure learner agency or autonomy (see Supplementary Information sections *Scale Extraction* and *Scale Item Preprocessing* for details), clustered them to reveal their semantic structure, and compared this structure with the results obtained from definitions. Clustering analysis of these items using the same pipeline as in the definition analysis yielded two major clusters (Figure 4a-b; see Supplementary Information section *Clustering Analysis of Scale Items* for details). Cluster S1 ($n$ = 2,183) centered on directing and monitoring learning tasks, and Cluster S2 ($n$ = 517) centered on inner volition and authenticity (Figure 4c-d).

From the perspective of the jingle-jangle fallacy, scale items exhibited a similar pattern. Agency and autonomy item embeddings showed no clear semantic boundary in the embedding space (Figure 4a). The two-cluster solution (S1/S2) yielded a substantially higher silhouette score (0.100) than the original autonomy-agency grouping (0.023), indicating that the autonomy-agency label provides even weaker separation than an unsupervised partition. This observation was further supported by pairwise cosine similarity analyses. Specifically, across all scale items, the mean similarity of agency-agency item pairs ($M$ = 0.298, $SD$ = 0.101) was even significantly lower than that of cross-construct autonomy-agency item pairs ($M$ = 0.309, $SD$ = 0.103; $\Delta = -0.011$, Cohen's $d$ = −0.110, permutation $p < .001$), which is a reversal of the expected pattern (see Supplementary Figure S6 for details). These results indicate that the jingle-jangle fallacy observed at the definitional level is even more pronounced in scale measurement of learner agency and autonomy.

When we computed pairwise cosine similarities between all scale-item embeddings and the definition embeddings and then averaged them by cluster and original label (Figure 4e), a relative diagonal correspondence emerged between item clusters and definition clusters. S1 showed highest correspondence with the C1 definition cluster (mean similarity values ranging from 0.231 to 0.282), whereas S2 aligned most closely with the C2 definition cluster (mean similarity values ranging from 0.199 to 0.238). Cross-cluster similarities were notably lower: S1-C2 similarities ranged from 0.163 to 0.182 and S2-C1 similarities from 0.158 to 0.188. Yet the weakest similarities of all were found in the C3 column. Both scale-item clusters showed consistently low correspondence with the C3 definition cluster, with values ranging from only 0.143 to 0.170, substantially below the diagonal S1-C1 and S2-C2 values. These results indicate a structured coverage gap between definitions and scale measurement. Existing self-report instruments map relatively well onto the semantic spaces of learning regulation and control (C1) and intrinsic motivation and internal decision (C2), but show weaker correspondence with the social and relational action (C3). Together, these findings indicate that current scale-based instruments exhibit limited content validity with respect to the full conceptual space of learner agency and autonomy, particularly the sociocultural-related dimension.

To examine whether this coverage gap reflects a broader methodological pattern, we analyzed the methodologies used across the 4,976 empirical articles from which definitions were extracted (see Supplementary Information section *Operationalization of Constructs* and Supplementary Figure S7 for the coding scheme and coding results). Overall, scale-based measurement was the most common approach (31.9%), followed by interviews or focus groups (25.4%). However, the pattern differed markedly between the two constructs. Autonomy research relied much more heavily on scales (45.1%) than on interviews or focus groups (16.9%) or text or document analysis (9.6%). Agency research, by contrast, showed a much more qualitative profile, with interviews or focus groups (38.0%), text or document analysis (24.8%), and observation or ethnography (20.4%) all more common than scale-based measurement (12.9%). A similar pattern appeared when articles were grouped by the definition dimension they contained. C1 and C2 definitions were more often associated with scale-based operationalization (40.8% and 32.6%, respectively), whereas C3 was predominantly associated with interviews or focus groups (41.2%), text or document analysis (25.2%), and observation or ethnography (23.3%), with scales accounting for only 8.2% of studies. These findings suggest that the weak representation of C3 definitions in the scale-item space aligns with a broader methodological pattern in this area of research, rather than reflecting only a gap in scale development. More broadly, C3 characterizes forms of action that are socioculturally mediated, temporally embedded, and relationally constituted, and these forms of action have more often been studied through qualitative methodologies than through scale instruments, possibly because qualitative approaches are better suited to capturing their contextual and dynamic nature.

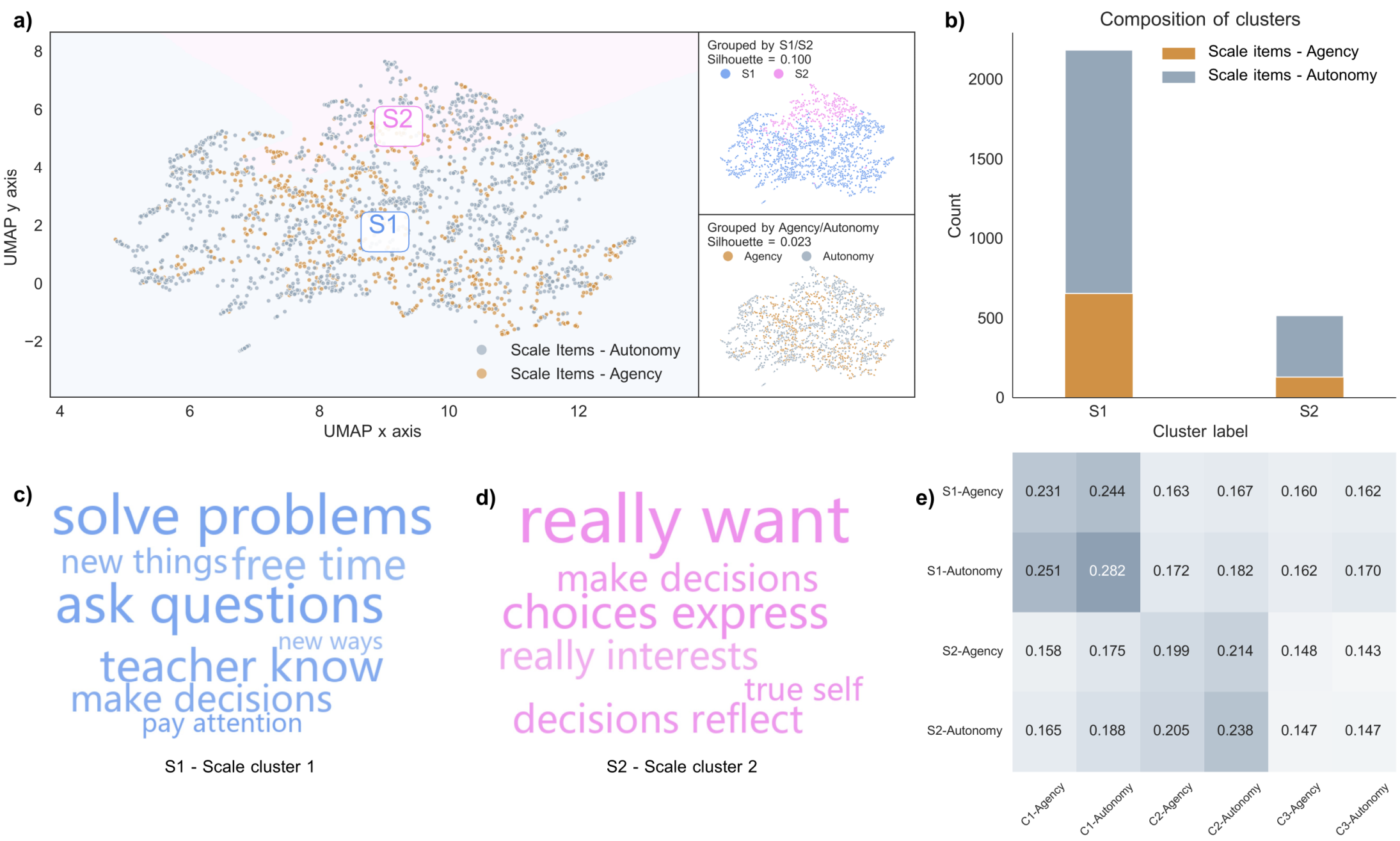

| | C1-Agency | C1-Autonomy | C2-Agency | C2-Autonomy | C3-Agency | C3-Autonomy |
|---|---|---|---|---|---|---|
| S1-Agency | 0.231 | 0.244 | 0.163 | 0.167 | 0.160 | 0.162 |
| S1-Autonomy | 0.251 | 0.282 | 0.172 | 0.182 | 0.162 | 0.170 |
| S2-Agency | 0.158 | 0.175 | 0.199 | 0.214 | 0.148 | 0.143 |
| S2-Autonomy | 0.165 | 0.188 | 0.205 | 0.238 | 0.147 | 0.147 |

**Figure 4. Semantic structure of scale-based measurement items. (a)** UMAP visualization of the semantic embedding space for the extracted scale items, grouped into two main clusters (S1 and S2). **(b)** Stacked bar chart showing the composition of the scale item clusters by construct (agency vs. autonomy). **(c-d)** Word clouds showing the most prominent terms in S1 and S2, respectively, based on c-TF-IDF scores. Common English function words were removed before term-frequency computation, and all displayed terms are 2-grams. **(e)** Heatmap of the mean pairwise cosine similarities between the four scale-item subgroups (S1-agency, S1-autonomy, S2-agency, and S2-autonomy) and the six definition-cluster subgroups (C1-, C2-, and C3-related agency and autonomy definitions).

### Generative AI Research Amplifies C1 Concentration

Given the rapid growth of research on learner autonomy and agency in GenAI-related learning contexts, we examined how definitions in this emerging literature are distributed across the three semantic clusters identified above, with the pre-2023 corpus serving as a baseline for comparison (see Supplementary Information section *Subsample Analysis of Generative AI Related Literature* for details). Using keyword-based matching of titles, abstracts, and author keywords, we identified 140 GenAI-related articles contributing 223 definitions, which were compared with the pre-2023 corpus. All matched articles were manually screened on the same fields to confirm that they genuinely discussed GenAI-related learning contexts. Subsequent analyses were conducted at the definition level. As shown in Figure 5, in terms of construct choice, GenAI-related articles showed only a slight shift toward agency relative to autonomy, with 50.7% autonomy and 49.3% agency, compared with 61.1% autonomy and 38.9% agency in the pre-2023 corpus. Although this difference was statistically significant ($\chi^2(1) = 9.39$, $p = .002$, Cramér's $V = 0.041$), the effect size was negligible. By contrast, the semantic distribution of definitions showed a more pronounced shift, as both agency and autonomy definitions in GenAI-related research were substantially more concentrated in C1 than those in the pre-2023 corpus. This shift was especially pronounced for agency, for which 55.4% of definitions in GenAI-related articles fell in C1, compared with 31.4% in the pre-2023 corpus, while the proportions in C2 and C3 declined from 28.7% to 17.3% and from 40.0% to 27.3%, respectively ($\chi^2(2) = 27.7$, $p < .001$, Cramér's $V = 0.113$). A parallel pattern was observed for autonomy, with 65.4% of definitions in GenAI-related articles falling in C1, compared with 50.3% in the pre-2023 corpus, alongside corresponding decreases in C2 from 43.5% to 32.7% and in C3 from 6.2% to 1.8% ($\chi^2(2) = 11.1$, $p = .004$, Cramér's $V = 0.058$). Taken together, these findings suggest that GenAI-related research does not substantially change which construct scholars adopt. Rather, it is associated with a consistent shift in how both agency and autonomy are defined, further concentrating them in the semantic domain of learning regulation and control.

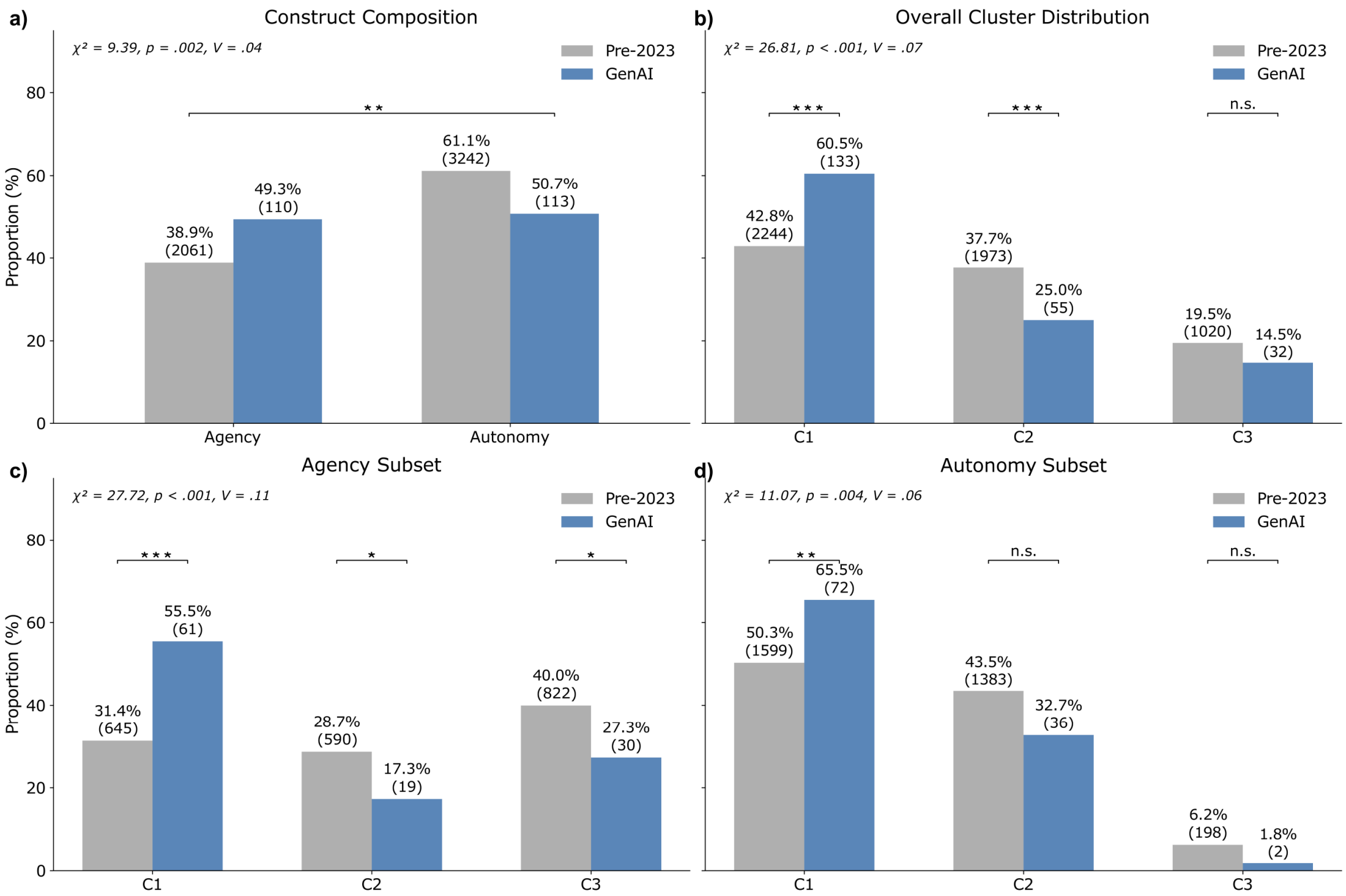

**Figure 5. GenAI-related research amplifies C1 concentration across both constructs.** Bars show the proportion (%) of definitions in the pre-2023 corpus (grey) and in GenAI-related articles (blue), with counts given in parentheses. **(a)** Construct composition of GenAI-related and pre-2023 definitions (agency vs. autonomy). **(b)** Distribution of all definitions across the three semantic clusters (C1 to C3). **(c-d)** Cluster distribution within agency definitions and autonomy definitions, respectively. The cluster analyses in (b-d) exclude definitions flagged as outliers in the clustering and not assigned to C1 through C3 (3 of 223 GenAI definitions and 66 of 5,303 pre-2023 definitions), so the cluster totals (220 GenAI and 5,237 pre-2023) are smaller than the construct totals in (a). Each panel subtitle reports the omnibus association ($\chi^2$, $p$, and Cramér's $V$). Brackets above the bars indicate statistical significance. In (a) the bracket reflects a single test on the two complementary

categories and is therefore uncorrected. In (b-d) the brackets reflect post-hoc cell-wise comparisons following a significant omnibus test, with $p$-values Bonferroni-corrected for the three clusters within each panel. * $p < .05$, ** $p < .01$, *** $p < .001$, n.s. = not significant.

## Discussion

By analyzing large-scale definitional data, we provide an empirical, bottom-up taxonomy of learner agency and autonomy. Our analysis shifts the focus from normative prescriptions (i.e., what these constructs should mean) to the descriptive reality of how they are operationalized in scientific discourse. The evidence reveals extensive jingle-jangle fallacies: learner agency and autonomy function as broad, umbrella-like constructs covering a shared conceptual space rather than marking two discrete theoretical territories. Data-driven clustering further identified three shared latent semantic dimensions: learning regulation and control (C1), intrinsic motivation and internal decision (C2), and social and relational action (C3). Existing scales, however, systematically neglect C3 and fail to discriminate between the two constructs within C1 and C2. Critically, current generative AI research in education concentrates on learning regulation and control, narrowing the behavioral repertoire that AI-mediated learning environments are designed to cultivate. Together, this three-dimensional structure functions not merely as a descriptive map but as a diagnostic framework for evaluating the conceptual coverage of measurement instruments and research. By providing empirical coordinates for this construct landscape, this study lays the groundwork for more precise theorizing, more valid measurement, and more productive interdisciplinary dialogue.

The identified three-dimensional structure is not arbitrary; each dimension carries a distinct theoretical lineage that explains the convergence of learner agency and autonomy. The C1 dimension is mainly composed of semantic units such as “taking charge of learning” and “learning process,” showing a clear task-oriented character. Its origin traces to Holec’s (1981) foundational work in adult education, which defined autonomy as the ability to take responsibility for one’s own learning, including different stages of learning-related decision-making. C1 therefore represents a structured framework of operational management. In contrast to the external focus of C1, C2 addresses internal psychological mechanisms. Its core vocabulary includes “making decisions” and “intrinsic motivation.” This dimension resonates with Little’s account of autonomy (1991) and Self-Determination Theory (Deci & Ryan, 1987, 2000), both of which emphasize perceived freedom, volition, and self-endorsement as basic psychological needs. C3 captures a shift from an individual-centered perspective to a relational perspective, highlighting semantic units such as “socioculturally mediated” and “temporally embedded.” This dimension is rooted in Ahearn’s work from the sociocultural perspective (Ahearn, 2001), where the capacity to act is understood as dynamically emerging within sociocultural networks. Collectively, the three dimensions provide an epistemological spectrum: from concrete task operation (C1), to internal motivation (C2), and then to broader social and relational networks (C3), showing why the two constructs, despite carrying different labels (learner agency and autonomy), end up describing much of the same conceptual ground. This convergence is not coincidental. Over the long development of the literature, both learner agency and autonomy were mobilized to address the same practical imperative: moving away from teacher-centered models toward learner self-direction (Benson, 2007; OECD, 2019). In doing so, both gradually expanded into umbrella-like labels absorbing operational, psychological, and social perspectives alike (Mercer, 2011; Murray, 2014). From a functionalist linguistic standpoint, the theoretical boundary between agency and autonomy, however rigorously articulated, never fully penetrated researchers' everyday discursive habits (Boleda, 2020). What the semantic landscape documents, then, is the accumulated weight of decades of collective usage driven by shared practical concerns rather than theoretical coordination.

Critically, shared semantic space does not imply construct interchangeability. Although learner agency and autonomy co-occupy the same semantic space, their distributions within each dimension are asymmetrical, revealing that their conceptual cores remain distinct. Within C1, autonomy definitions are more densely concentrated, forming an autonomy-dominated discursive territory, while agency definitions in this region are pulled toward autonomy's operational vocabulary, suggesting that when researchers discuss agency in terms of task management and learning control, they draw on a language that is effectively autonomy's own. The mirror pattern holds in C3: this dimension is predominantly anchored by learner agency, while autonomy definitions fail to form any coherent aggregation, remaining scattered at the periphery. C2 occupies a more intermediate position, where both constructs retain distinguishable internal semantic structures within a shared zone of conceptual mixture. Taken together, this asymmetrical mapping points to a clearer picture: autonomy's most distinctive conceptual territory lies in C1, while agency's lies in C3; and C2 represents the psychological common ground where the two most closely overlap. The semantic landscape is therefore not a uniform mixture of two interchangeable labels, but a structured terrain of construct-specific territories and a shared intermediate zone. The three-dimensional classification proposed here offers a more productive unit of analysis than construct labels alone. Rather than asking whether learner agency and autonomy are the same or different, future research can ask a more precise question: within which specific dimension do the two constructs diverge, and to what degree? Density distributions across dimensions, rather than label assignment, may prove the more reliable basis for both theoretical distinction and empirical operationalization (Larsen & Bong, 2016).

Conceptual jingle-jangle does not stay confined to theory, but cascades into measurement. When scales are developed under construct labels rather than construct content, they inherit the same conceptual confusion from which they were built: the boundaries drawn in measurement reflect the boundaries, or lack thereof, drawn in theory. Indeed, existing instruments for learner agency and autonomy reveal systematic content validity deficits that directly mirror the conceptual confusions identified above. Most critically, current scales largely overlook C3: scale items are predominantly aligned with C1 and C2, while their similarity with C3 remains consistently low. This is not entirely surprising. Studies located in the C3 cluster rely more heavily on qualitative methods, reflecting a broader epistemological conviction that socioculturally mediated and dynamically interactive forms of relational action require context-sensitive approaches that preserve ecological validity (Brewer & Crano, 2024; Fenwick et al., 2015; Greeno, 1998; Goffman, 1983; Lincoln & Guba, 1985). Traditional scales, which ask participants to engage in retrospective introspection outside concrete contexts, carry inherent epistemological limits for capturing such phenomena (Shiffman et al., 2008). Yet individual methodological preferences do not eliminate a field-level problem: without quantitative tools for C3, an entire dimension of learner agency and autonomy remains invisible to cumulative, cross-study synthesis, and systematically underrepresented from the evidence base that informs theory-building and intervention design.

The measurement challenges within C1 and C2 are equally serious, if less visible. Learner agency scale items often show greater semantic similarity to autonomy constructs than to themselves, indicating that current measurement practices have been organized around linguistic labels rather than conceptual content. The result is not measurement of two distinct constructs, but the same semantic territory surveyed twice under different names. The measurement of agency, in particular, requires fundamental reconsideration. Within C1 and C2, agency scales do not form a structure clearly distinct from autonomy scales, raising questions about their discriminant validity that incremental scale refinement alone cannot resolve. Compounding this, C3, the sociocultural dimension most distinctively anchored by agency, is precisely the one that existing quantitative instruments fail to reach, meaning that what is most characteristic of agency is also what is least measured. The two analyses are not independent. The measurement deficits identified in the scale analysis are not coincidental; they are the operational residue of the conceptual confusion documented in the definitional analysis. Scales developed under construct labels rather than construct content inherit the same jingle-jangle structure: they fail to discriminate between agency and autonomy where the two conceptually diverge, and fail to reach the sociocultural dimension (C3) that is most distinctively associated with agency. This cascade from conceptual to operational level represents a content validity problem in the strict sense: existing instruments do not measure what researchers claim to measure, because what researchers claim to measure has never been clearly defined.

Addressing these deficits requires dimension-specific strategies. For C1 and C2, scale development must be rebuilt from the three semantic dimensions identified in this study, treating each dimension as the primary unit of operationalization rather than the construct label. Items should be selected and validated on the basis of their semantic alignment with a specific dimension, not their surface resemblance to existing agency or autonomy scales. For C3, where traditional self-report approaches carry inherent epistemological limits, multimodal digital traces that record learners' authentic interaction trajectories in non-intrusive environments offer a more promising pathway for capturing emergent relational action without sacrificing ecological validity (Dowell et al., 2019). Only by anchoring measurement to what researchers actually study, rather than what they call it, can the field move toward instruments that genuinely distinguish these constructs and accumulate comparable evidence across studies.

Finally, the stakes of this clarification extend beyond academic taxonomy and carry direct practical consequences, nowhere more urgently than in the context of generative AI. The dimensional analysis of generative AI-related publications reveals a marked concentration in C1, with both learner agency and autonomy definitions predominantly instantiating the regulation-and-control dimension relative to the broader corpus, a pattern consistent with prior empirical observations in the field (Achuthan, 2025; Zai & Zhou, 2026). This concentration appears to operate through two complementary pathways: a portion reflects definitions that explicitly frame learner autonomy and agency as regulatory control over AI tools; the remainder persists even among definitions that make no explicit reference to AI tools, suggesting a broader contextual pull toward regulation-and-control framings that warrants further investigation. Critically, this C1 concentration is not occurring at the level of which construct researchers claim to study, but at the level of which conceptual dimensions they are actually instantiating. Only by moving beneath construct labels to definitional substance, precisely what the present framework enables, does this structural skew become detectable. As generative AI systems become more deeply embedded in educational infrastructure, the conceptual foundations informing their design matter. However, researchers may not recognize that C1 represents only one dimension of what autonomy and agency theoretically encompass: if the volitional and sociocultural dimensions addressed by C2 and C3 remain structurally underrepresented in the accumulating evidence base, the conceptual resources available to guide system design will be systematically incomplete, with implications for how learner agency and autonomy are understood, measured, and ultimately supported in AI-mediated learning contexts.

Beyond the substantive findings on learner agency and autonomy, the present study offers a replicable workflow that can be extended to other constructs of interest. Therefore, a question worth addressing is what the semantic landscape recovered through this pipeline truly reflects. We argue that it reflects collective conceptual understanding. The corpus consists exclusively of explicit academic definitions, not casual discourse. When a scholar formally defines a construct in a publication, that definition is their conceptual understanding: it represents a deliberate, considered articulation of what the construct means. The corpus of explicit academic definitions therefore provides a valid window into how individual researchers conceptualize the constructs in question. Yet the semantic landscape that emerges from aggregating thousands of such definitions carries a deeper epistemic significance. It is not merely a description of how researchers happen to use these terms, but a reflection of the latent conceptual structures that have stabilized across authors, time periods, and theoretical traditions. From the perspective of linguistic functionalism, meaning is not a property of individual utterances but is constituted through patterns of use within a community of practice (Wittgenstein, 1953; Searle, 1969). Each definition is simultaneously an act of positioning and an act of contribution: it draws on existing shared understandings while also, incrementally, reshaping them. In this sense, the semantic landscape we recover does not merely describe the sum of individual authorial choices; it reveals the deeper epistemic architecture through which the field collectively understands learner agency and autonomy in research practice. What this pipeline cannot resolve, however, is the normative question of how the constructs should be distinguished going forward. The empirical coordinates we provide describe the accumulated state of the field; translating them into prescriptive conceptual boundaries will require theoretically grounded consensus-building that computational mapping alone cannot substitute. Yet this is precisely where the value of the computational approach lies: because it offers neutral empirical coordinates rather than a theoretical stance, it can bring researchers from diverse academic traditions onto a common map. The goal is not to terminate theoretical innovation or erase minority perspectives, but to establish a foundational consensus from which diverse academic communities can move past endless terminological disputes toward cumulative knowledge-building.

Nevertheless, there are other concerns that must be acknowledged when applying LLM-based semantic mapping to conceptual analysis. A first concern is whether semantic similarity in high-dimensional space adequately captures the complexity of these constructs. We addressed this directly by validating embedding-derived similarity against empirically reported Cronbach’s alpha coefficients, finding that distributional patterns in semantic space meaningfully reflect the internal consistency structure of existing scales. This does not mean that semantic similarity captures everything, but it does suggest that it captures a psychologically meaningful portion of constructs. A second concern is representational: clustering algorithms inherently maximize within-cluster cohesion, privileging majority patterns and potentially obscuring minority or emergent conceptualizations located in the statistical long tail. Highly innovative, context-specific insights may be underrepresented in the resulting landscape. The textual convergence we observe therefore reflects the dominant academic discourse, a “linguistic reality” rather than the full range of possible meaning-making (Bender et al., 2021). A third concern is cultural: all analyses were conducted on English-language corpora, yet agency and autonomy are not cross-culturally invariant. Western traditions tend to emphasize independence-oriented conceptions of these constructs, whereas other cultural contexts foreground more relational framings (Chirkov et al., 2003; Yang et al., 2023). Future work should therefore extend this approach to broader multilingual corpora and test it across different large language models, so as to mitigate the cultural biases that any single model may inherit during pre-training. Finally, our analysis relies on LLM-assisted extraction at scale. Recent work shows that LLMs can match human coders in accuracy, making them viable for large-scale text processing (Dagdelen et al., 2024; Gilardi et al., 2023; Rathje et al., 2024). We further strengthened reliability through expert review and iterative refinement of prompts beforehand, and verbatim checks confirming fidelity to the original texts afterward, finding that extractions were largely consistent with human judgment. Still, LLMs remain prone to hallucination and data noise, and a small proportion of extractions may diverge from human judgment. Narrowing this gap remains an ongoing challenge for the field.

# Methods

**Constructing a Large-Scale Definition Corpus on Learner Agency and Autonomy**

We first constructed a large-scale corpus of literature on agency and autonomy in educational contexts using systematic database search, metadata screening, and full-text acquisition. We searched Web of Science with *TS=("autonomy" OR "agency") AND TS=(educat* OR learn*)* and Scopus with *TITLE-ABS-KEY(("autonomy" OR "agency") AND (educat* OR learn*))*, yielding 197,870 metadata records in total. Records were filtered to retain only non-duplicate entries with a valid DOI and at least one occurrence of agency or autonomy in the title, abstract, or keywords. We then used AI-assisted metadata screening (GPT-4o) to retain studies examining agency or autonomy in learning contexts involving learners or students. Full texts were obtained through open-access retrieval, Text and Data Mining (TDM) through APIs from publishers, and targeted manual collection. The final corpus comprised 14,611 articles, providing broad coverage of accessible research on agency and autonomy. Details of the reference retrieval procedures are provided in the Supplementary Information section *Literature Corpus Construction*.

Using this full-text corpus, we extracted explicit definitions concerning agency and autonomy via the *GLM-4.7* model. Extraction was restricted to definitions pertaining to learners or students in learning settings and excluded uses referring to agency or autonomy in other actors or contexts, including those related to teachers, organizations, politics, machines, and systems. Definitions were retained in the original wording of the source, with only introductory framing phrases (e.g., “X is defined as”) removed. This procedure yielded 9,657 definitions from 5,388 articles. We then identified the language of each entry and retained 8,954 English-language definitions from 4,976 articles for subsequent analyses, comprising 5,272 definitions of autonomy and 3,682 definitions of agency. Details of the extraction procedure are provided in the Supplementary Information section *Definition Extraction*.

**Semantic Structure Analysis of Agency- and Autonomy-Related Definitions Using Embeddings**

To analyze the semantic structure of agency- and autonomy-related definitions, the retained English-language definitions were represented using embeddings generated by OpenAI’s *text-embedding-*

*3-large* model. Each definition text was transformed into a 3,072-dimensional vector. This model was selected because it provides high-capacity semantic representations of text data and has been validated in recent research on semantic text analysis (Huang et al., 2025; Wulff & Mata, 2025b). We then used these embedding vectors to identify core semantic clusters among the definitions, following the general embedding-based analytic logic exemplified by BERTopic (Grootendorst, 2022), with a pipeline of dimensionality reduction and clustering, followed by keyword extraction and visualization (reported in Results). The embeddings were first reduced using UMAP and then clustered using hierarchical or K-means algorithms across multiple hyperparameter settings. We varied the UMAP neighborhood size, output dimensionality, and minimum distance, as well as the number of clusters for both agglomerative hierarchical clustering and K-means, and the minimum cluster-size criterion used to retain substantive clusters in the hierarchical solutions. Candidate solutions were compared using silhouette scores (Rousseeuw, 1987) computed in the original embedding space. The optimal solution was a three-cluster hierarchical partition. Based on cluster keywords and representative definitions, the three clusters were labeled C1: learning regulation and control ($n$ = 3,873; agency = 1,255, autonomy = 2,618), C2: intrinsic motivation and internal decision ($n$ = 3,250; agency = 1,000, autonomy = 2,250), C3: social and relational action (n = 1,688; agency = 1,419, autonomy = 269), and an outlier cluster ($n$ = 143; agency = 8, autonomy = 135). Details of the embedding computation and clustering are provided in the Supplementary Information sections *Embedding Computation and Validation of Definitions* and *Clustering Analysis of Definitions*.

**Semantic Analysis of Psychometric Scale Items Measuring Agency and Autonomy**

After analyzing the conceptual structure of agency- and autonomy-related definitions, we then examined whether psychometric scales in the existing literature could adequately represent these semantic dimensions. To achieve comprehensive coverage of the scale-development literature, we scanned the full literature corpus to identify all source articles linked to measurement instruments and retrieved their full texts, supplemented by manual searches for recently developed instruments, yielding 905 candidate papers in total. Eligible papers were required to provide complete scale items in English, and only scales or subscales whose names explicitly contained autonomy- or agency-related terms were retained. The final scale corpus contained 2,700 items drawn from 268 articles. Of these items, 1,916 measured autonomy and 784 measured agency. Before computing embeddings, the wording of all extracted scale items was standardized to reduce superficial linguistic variation while preserving substantive item meaning. Using the same analytic approach as in the definition analysis, embeddings for all standardized scale items were computed with OpenAI's *text-embedding-3-large* model and subjected to clustering analysis using the same procedure and hyperparameter search space. The best-performing solution was a two-cluster hierarchical clustering solution, labeled S1 and S2. S1 included 2,183 items (autonomy = 1,528, agency = 655), whereas S2 included 517 items (autonomy = 388, agency = 129). Details of the scale extraction procedure, item preprocessing, embedding validation, and clustering analysis are provided in the Supplementary Information sections *Scale Extraction*, *Scale Item Preprocessing*, *Embedding Computation and Validation of Scale Items*, and *Clustering Analysis of Scale Items*.

Beyond analyzing psychometric scales, an LLM-assisted full-text scan was used to examine how learner agency and autonomy had been operationalized across the broader literature. This provided a more comprehensive picture of operationalization practices and complemented the scale-based findings. Operationalizations were coded into six broad categories, namely scale-based quantitative measures, behavioral quantitative indicators, experimental or intervention-based operationalizations, interview- or focus-group-based qualitative approaches, observation- or ethnography-based qualitative approaches, and text- or document-based qualitative approaches. Across all articles combined, scale-based quantitative measures and interview- or focus-group-based qualitative approaches were the two most common operationalization types, corresponding to 31.9% and 25.4% of articles, respectively. Details of the coding framework and procedure are provided in the Supplementary Information section *Operationalization of Constructs*.

**Validation of the LLM-Based Analysis**

Because several key steps in this study relied on LLM-based analysis, multiple validation analyses were conducted (see Supplementary Information for full details). First, extraction fidelity for both definitions and scale items was evaluated through exact string matching with the source documents. Verbatim matches were observed for 95.5% of the 8,954 definitions and 95.7% of the 2,700 scale items. The remaining cases involved only minor formatting differences and no substantive alteration of meaning. Accordingly, all AI-extracted definitions and scale items were retained for analysis.

Second, we examined whether the embedding model used here converged with alternative embedding models in representing the semantic structure of agency- and autonomy-related constructs. Embeddings were also generated using alternative models from Qwen and ZhipuAI, and the resulting definition-by-definition cosine similarity matrices were compared with those derived from OpenAI embeddings. The similarity matrix derived from the OpenAI embeddings showed high correspondence with those produced by Qwen (Spearman's $\rho$ = 0.836) and ZhipuAI (Spearman's $\rho = 0.757$), indicating strong cross-model consistency. These results support the use of the *text-embedding-3-large* model for representing the semantic structure of agency- and autonomy-related definitions.

Third, additional evidence for embedding validity was drawn from the psychometric properties available in the scale corpus. Specifically, embedding-based estimates of subscale internal consistency were compared with empirically reported Cronbach's α values. The estimated and reported α values were positively correlated for both autonomy subscales ($r = 0.271$, $p = .003$, $n = 116$) and agency subscales ($r = 0.355$, $p = .002$, $n = 71$). Item-level structural separation was also examined by testing whether items within the same subscale were more similar to each other than to items from other subscales. The results showed that 68.8% of autonomy subscales and 82.9% of agency subscales displayed clear structural separation in embedding space ($z$-score > 2). Together, these findings support the use of embeddings to represent the semantic structure of agency- and autonomy-related measurements.

Fourth, the robustness of the clustering results was examined within the definition corpus. The clustering analysis was repeated separately for the deduplicated autonomy-related and agency-related definitions using the same procedure and hyperparameter search space described above. Each subsample-specific solution was then compared with the corresponding subset of the full-sample solution using normalized mutual information (NMI) and the adjusted Rand index (ARI). Agreement between the subsample and full-sample solutions was substantial for both autonomy (n = 5,015; NMI = 0.671, ARI = 0.757) and agency (n = 3,585; NMI = 0.604, ARI = 0.610). Taken together, these results indicate that the full-sample clustering captures construct-specific semantic structure that is reproducible within both domains.

**Acknowledgements.** This study was supported by the National Natural Science Foundation of China (62577038) and Shenzhen Science and Technology Program (AI2026029). The authors would like to thank Ms. Xinmiao Zhang for her valuable suggestions.

# Supplementary Information for Large-scale semantic mapping of learner agency and autonomy reveals what measurement and generative AI research overlook

## Literature Corpus Construction

To assemble the literature corpus, we conducted systematic searches in Web of Science and Scopus using database-specific boolean queries on January 22, 2026. The query used in Web of Science was *TS=("autonomy" OR "agency") AND TS=(educat* OR learn*)*, and the query used in Scopus was *TITLE-ABS-KEY(("autonomy" OR "agency") AND (educat* OR learn*))*. These searches returned 197,870 records in total, comprising 59,755 from Web of Science and 138,115 from Scopus. Within the initial retrieval set, 76,585 records mentioned only *agency*, 70,732 mentioned only *autonomy*, 2,684 mentioned both terms, and 47,869 mentioned neither term in the title, abstract, or keywords.

We then applied a metadata-based filtering step. Records were retained only if they had a valid DOI, were non-duplicate at both the title and DOI levels, and contained at least one occurrence of agency or autonomy in the title, abstract, or keywords. After this step, 85,392 records were retained. Of these, 43,394 mentioned only agency, 40,412 mentioned only autonomy, and 1,586 mentioned both terms.

We next carried out AI-assisted screening based on bibliographic metadata, including titles, abstracts, and keywords, using OpenAI GPT-4o. Each record was evaluated against two criteria: the primary setting had to be a learning context, including formal schooling, online learning, or informal education, and the focal population had to include learners or students. Records that failed to satisfy either criterion were excluded. This screening yielded 26,335 records. Among these, 11,521 mentioned only *agency*, 14,206 mentioned only *autonomy*, and 608 mentioned both terms in the title, abstract, or keywords. The reliability of AI-assisted screening was evaluated using a stratified random sample of 200 records, consisting of 100 records accepted by the model and 100 records rejected by it. These records were randomized and independently reviewed by a trained human expert in the field of education who had been instructed on the screening criteria but was blind to the AI decisions during this review. Agreement between AI and human judgments yielded Cohen's $\kappa = 0.65$, with a sensitivity of 0.874 and a specificity of 0.788, indicating that most records judged relevant by the human reviewer were successfully retained. Because the screening decision was based only on bibliographic metadata rather than full texts, this level of agreement was considered acceptable for large-scale corpus construction.

Full texts were then obtained through three complementary strategies. For records identified as open access, PDFs were retrieved through the Unpaywall (Piwowar et al., 2018) and OpenAlex (Priem et al., 2022) APIs, with Zotero's full-text retrieval function used to recover PDFs not obtained through these APIs (Corporation for Digital Scholarship, 2024). For non-open-access records, we first checked whether the DOI was associated with text and data mining authorization. When such authorization was confirmed, we retrieved the content through publisher-provided access channels (Elsevier, 2024; Springer Nature, 2024; Crossref, 2024), using API-based download when available and manual retrieval when necessary. Records without confirmed text and data mining authorization were excluded from bulk downloading in order to comply with copyright restrictions. To reduce the possibility of omitting highly relevant studies, we also manually retrieved and supplemented the top 1,000 most relevant records from each database (Web of Science and Scopus) under institutional subscription access. Across these procedures, full texts were successfully obtained for 14,611 of the 26,335 records (55.5%). Within the resulting full-text corpus, 5,667 records mentioned only *agency*, 8,527 mentioned only *autonomy*, and 417 mentioned both terms.

## Definition Extraction

From the full-text corpus, we extracted explicit definitional statements concerning *agency and autonomy* in learner- or student-centered educational contexts. We used GLM-4.7, a large language model that supports long-context input and native PDF processing (Glm et al., 2024), to read each full text and identify definitional statements. The extraction prompt instructed the model to retain only definitions pertaining to learners or students in learning settings and to exclude definitions referring to agency or autonomy in other actors or contexts, including teachers, organizations, politics, machines, and systems.

For each extracted entry, we retained three fields: the defined term, the original definition text, and its classification as agency or autonomy. Classification was based on the defined term itself: terms containing *autonomy* or *autonomous* were classified as autonomy, whereas terms containing *agency* or *agentic* were classified as agency. Definitions were retained in the original wording of the source text, without paraphrasing, inference, or summarization. Introductory definitional phrases, such as “is defined as,” “refers to,” and “is understood as,” were removed.

This procedure yielded 9,657 definitional statements from 5,388 articles, including 5,781 classified as *autonomy* and 3,876 classified as *agency*. Because extraction was conducted without restricting source language, we then identified the language of each extracted entry using an LLM prompted to return ISO 639-2 three-letter language codes only. Of the 9,657 extracted definitions, 8,954 were identified as English. The remaining entries were distributed across Spanish (296), Portuguese (149), Russian (97), French (68), German (17), Korean (15), Norwegian (14), Serbian (9), Japanese (9), Turkish (6), and several other languages with fewer than five entries each. Subsequent analyses were restricted to the 8,954 English-language definitions, which were drawn from 4,976 articles and included 5,272 definitions classified as *autonomy* and 3,682 classified as *agency*. Among these 4,976 articles, the mean number of definitions extracted per article was 1.80 (*SD* = 1.46). A total of 3,018 articles yielded a single definition, 1,078 yielded two, and the maximum for any single article was 17. Overall, 82.3% of articles contributed no more than two definitions, and the ten highest-contributing articles accounted for only 1.5% of all definitions, indicating that the extraction was distributed relatively evenly across the corpus rather than concentrated in a small number of sources.

We evaluated extraction quality of definitions through two complementary checks, an automated assessment of extraction fidelity and a manual expert review of a random sample. First, to assess extraction fidelity, we performed exact string matching between each extracted English-language definition and its source document. Of the 8,954 English-language definitions, 95.5% matched the source text verbatim. Manual inspection of the remaining 4.5% showed that most discrepancies reflected only minor formatting normalization, such as the removal of in-text citations and reference markers, with no substantive alteration of meaning. Only 23 entries (0.3%) could not be matched to any corresponding passage in the source document, suggesting possible hallucination by the language model. Second, a random sample of 200 extracted definitions was independently reviewed by one expert in the field of education, who evaluated each definition against the original source text and the predefined extraction criteria. The manual review confirmed that 182 (91.0%) fully met the predefined extraction criteria. The remaining cases pertained to constructs outside the target scope, such as teacher agency and civil agency, or were descriptive statements from the source text that the language model had incorrectly extracted as formal definitions. We therefore consider the LLM-based extraction to be acceptable for the analyses reported below.

We also conducted a contextual consistency check for agency-autonomy labels. Specifically, we searched the 100 characters preceding the best-matching span in the source text for the stems "autonom*" and "agen*" and evaluated whether detected stems were consistent with the assigned label. Among 8,954 definitions, 5,883 (65.7%) showed a corresponding stem (i.e., "agen*" for agency and "autonom*" for autonomy), 33 (0.4%) showed a reverse stem pattern, 129 (1.4%) contained both stems, and 2,909 (32.5%) contained neither stem in the 100 characters preceding the matched span. Thus, direct contradictory evidence was rare. The "neither" cases can be partly explained by the fact that, in some source texts, the terms may appear in abbreviated forms or at a greater distance from the definition. Overall, these results suggest limited agency-autonomy confusion in cases with identifiable local stem evidence.

## Spatiotemporal Distribution of Definitions

Following the extraction of definitional statements, we first examined the distribution of agency and autonomy definitions from temporal and spatial perspectives. The publication year of the source article for each of the 8,954 definitions was extracted from the metadata provided by Web of Science and Scopus. Institutional affiliation information was available for 99.1% of the source articles. We parsed the affiliation strings to identify the unique institutions represented in each study and then queried the Elsevier and OpenAlex APIs to retrieve the corresponding country or region and continent information, successfully obtaining geographic data for 98.3% of the institutions. All distributions reported below were computed at the definition level (i.e., each of the 8,954 definitions served as the unit of analysis). For institutional distributions, each contributing institution of a given definition was counted once. For continental distributions, each definition was attributed across continents using fractional weighting, with a definition involving n contributing countries assigned 1/n to each corresponding continent.

As shown in Supplementary Figure S1, both autonomy- and agency-related definitions were concentrated in recent decades. The most frequently contributing institutions differed between the two sets of definitions. Agency-related definitions were more frequently associated with the University of California System, University of Oslo, University of Helsinki, and Deakin University, whereas autonomy-related definitions were more frequently linked to Ghent University, University of Hong Kong, the Chinese University of Hong Kong, and Universidad de Extremadura. Geographically, agency definitions were concentrated in Europe (41.2%) and North America (29.7%), with a smaller share from Asia (16.4%), whereas autonomy definitions were most strongly represented in Asia (38.5%) and Europe (33.0%), followed by North America (19.2%). These patterns suggest that, despite a shared overall growth in recent decades, the two constructs differ in their institutional and geographic bases.

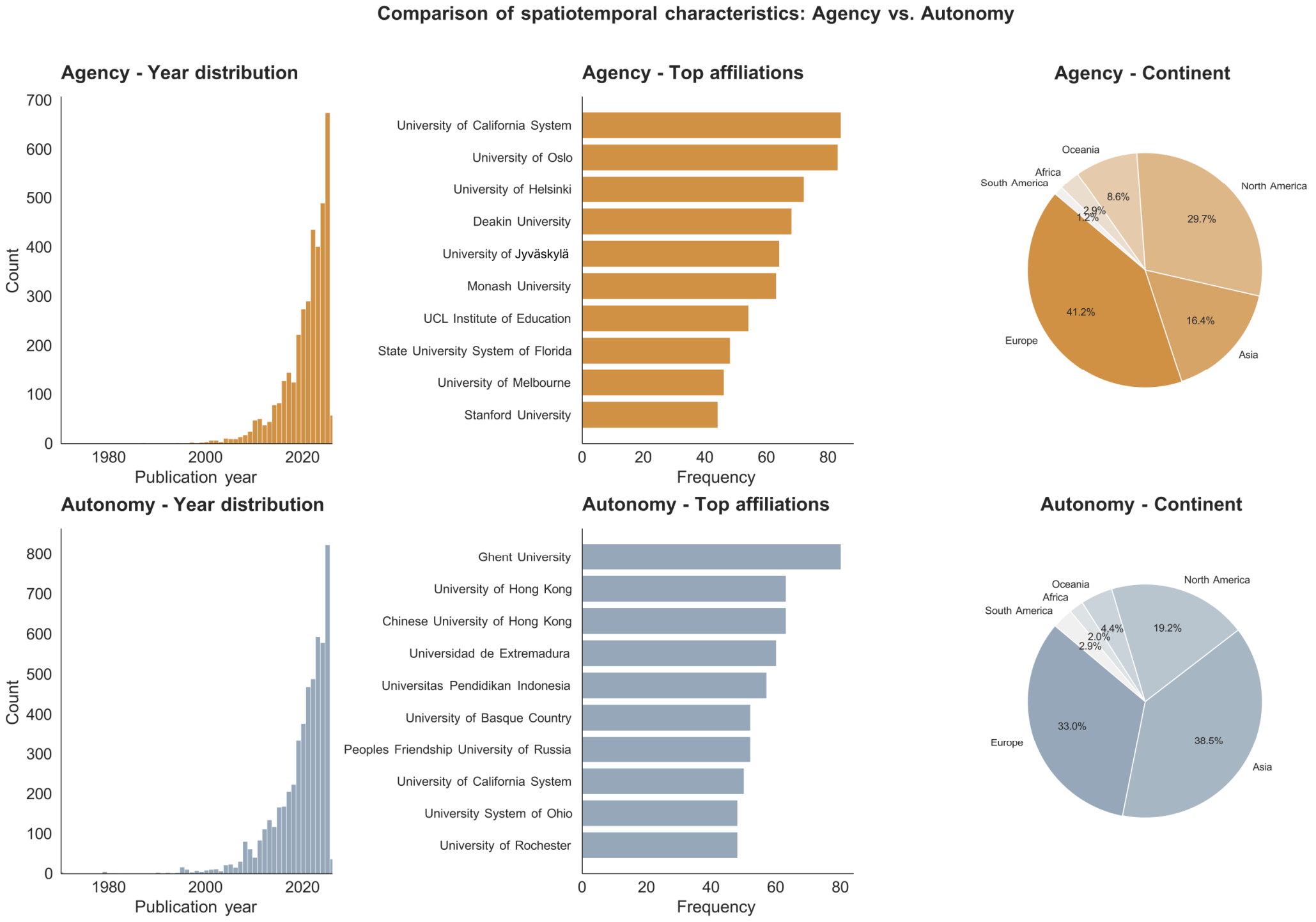

**Supplementary Figure S1.** Descriptive spatiotemporal characteristics of learner agency and autonomy definitions.

## Embedding Computation and Validation of Definitions

We computed semantic embeddings for the retained English-language definitions using OpenAI's *text-embedding-3-large* model, which produces 3,072-dimensional vectors. The subsequent semantic analyses were built upon computations derived from these vectors, with semantic similarity serving as the core measure. The similarity between any two definitions can be quantified by the cosine of the angle between their embedding vectors, defined as:

$$Sim(m,n) = \frac{E_m \cdot E_n}{||E_m||\,||E_n||}$$

where $E_m$ and $E_n$ are the embedding vectors of two definitions. Across the 8,954 definitions, the average pairwise cosine similarity was 0.364 (*SD* = 0.119).

We validated the embeddings from two perspectives, namely whether the pairwise cosine similarity values fell within a reasonable range, and whether the similarity structure was stable across different embedding models. To address the first, we established a reference baseline by computing embeddings for definitions from the ERIC Thesaurus (Institute of Education Sciences, n.d.), a comprehensive controlled vocabulary covering the full breadth of the education domain. Using the same OpenAI embedding model, the average pairwise cosine similarity among the 3,650 general educational definitions was 0.191 (*SD* = 0.087). The ERIC Thesaurus also provides curated links between related concepts. For the 17,316 related-concept pairs provided, the mean cosine similarity was 0.429 (*SD* = 0.131). The mean similarity of our definition corpus (0.364) fell between the general baseline (0.191) and the related-concept pairs (0.429). This ordering is consistent with expectations, as the definitions of agency and autonomy are expected to exhibit semantic overlap given their conceptual proximity while also encompassing distinct semantic components that reflect different theoretical perspectives.

To address the second, we generated alternative embeddings for the same 8,954 definitions using ZhipuAI's *embedding-3* model (2,048 dimensions) and Qwen's *text-embedding-v3* model (1,024 dimensions). For each model, we computed a definition-by-definition cosine similarity matrix and extracted the upper triangle. Spearman correlations between these vectors were then calculated to quantify the correspondence across models. The similarity structure derived from the OpenAI embeddings correlated at $\rho = .836$ with that derived from Qwen and at $\rho = .757$ with that derived from ZhipuAI. These results indicate that the semantic structure captured by the OpenAI embeddings was broadly consistent with that recovered by alternative models, despite differences in model provider and embedding dimensionality.

## Clustering Analysis of Definitions

Following the embedding-based text analytic logic exemplified by BERTopic, we analyzed the semantic structure of autonomy- and agency-related definitions using a pipeline of dimensionality reduction, clustering, and model selection. Clustering was performed on the 8,954 English-language definitions with valid embeddings generated by the *text-embedding-3-large* model. To reduce the influence of exact textual duplicates, we first deduplicated entries by definition text prior to hyperparameter tuning and clustering, yielding 8,594 unique definitions. Cluster assignments obtained on the deduplicated set were then mapped back to the full set of 8,954 records for descriptive reporting. The most frequent autonomy-related definition was "the ability to take charge of one's own learning" ($n$ = 61), and the most frequent agency-related definition was "the socioculturally mediated capacity to act" ($n$ = 19).

Because the embedding vectors were high-dimensional (3,072 dimensions), directly clustering them may hinder the recovery of meaningful semantic structure because of the challenges associated with high-dimensional representations. We therefore first applied Uniform Manifold Approximation and Projection (UMAP), a commonly used manifold-learning and dimensionality-reduction method in embedding-based clustering pipelines, before clustering. UMAP was fit with cosine distance and a fixed random seed. Hyperparameters were varied over n_neighbors = {10, 30, 50}, n_components = {50, 100, 150, 200}, and min_dist = {0.0, 0.05}, yielding 24 dimensionality-reduction configurations. We then evaluated two commonly used clustering algorithms on the reduced representations: agglomerative hierarchical clustering and K-means. For hierarchical clustering, we used average linkage and varied the initial number of clusters over k = {2, 3, 4, 5, 6, 7, 8}. To avoid overinterpreting very small residual groups, we additionally imposed a minimum cluster-size criterion such that only clusters containing at least a specified proportion of the corpus were retained as valid substantive clusters, with the minimum cluster ratio varied over {0.05, 0.03}. For K-means, the number of clusters was likewise varied over k = {2, 3, 4, 5, 6, 7, 8}. For each combination of UMAP and clustering hyperparameters, candidate solutions were evaluated using silhouette scores for valid clusters computed in the original embedding space with cosine distance. In total, we evaluated 336 hierarchical solutions and 168 K-means solutions.

The best-performing solution, and the one retained for interpretation in the main text, used

UMAP with n_neighbors = 50, n_components = 200, and min_dist = 0.0, followed by agglomerative hierarchical clustering with average linkage, an initial target of k = 4, and a minimum cluster ratio of 0.05. Under this criterion, only clusters containing at least 5% of the corpus were treated as valid substantive clusters. One of the four initial clusters fell below this threshold ($n$ = 143) and was therefore treated as an outlier cluster, leaving three substantive clusters in the retained solution. Inspection of the definitions in this small excluded cluster confirmed that they reflected outlier content rather than the main semantic structure of autonomy- and agency-related definitions. This configuration achieved the highest silhouette score in the original embedding space among the hierarchical candidates (silhouette = 0.113). The best K-means solution used UMAP with n_neighbors = 50, n_components = 200, and min_dist = 0.0, followed by K-means with k = 2 (silhouette = 0.109), but yielded a coarser and less substantively informative partition. We therefore retained the three-cluster hierarchical solution. In the mapped full corpus ($N$ = 8,954), C1 (learning regulation and control), C2 (intrinsic motivation and internal decision), and C3 (social and relational action) contained 3,873, 3,250, and 1,688 definitions, respectively, with one outlier cluster of 143 definitions.

Because silhouette scores are computed after excluding outliers, different hyperparameter configurations may operate on slightly different sample sizes depending on how many observations are flagged as outliers, potentially introducing a confounding factor into the model selection process. To validate the robustness of the selected hyperparameters, we identified the top 50 configurations for each method by silhouette score and took the union of all observations flagged as outliers across these solutions. Because K-means does not produce outlier-labeled observations, the union was contributed entirely by the top 50 hierarchical configurations, yielding 245 unique observations (2.7% of the total 8,954 observations). We then uniformly removed this outlier union from all configurations before recomputing silhouette scores, ensuring that every configuration was evaluated on an identical sample. Under this standardized comparison, the best hierarchical configuration retained its top rank among all 336 hierarchical solutions both before (silhouette = 0.113) and after (silhouette = 0.114) uniform outlier removal, confirming that the hyperparameter selection was not an artifact of differential outlier exclusion. Among the top 50 K-means configurations (out of 168 total), however, 9 solutions yielded higher silhouette scores than the best hierarchical configuration under this outlier-removed standardized comparison. Inspection of the cross-tabulations revealed that all 9 solutions used k = 2, collapsing the three-cluster hierarchical solution into two clusters. Approximately 80% of C3 definitions were merged into C2, with the remaining ~20% absorbed into C1, while the composition of C1 and C2 themselves remained largely unchanged. The marginal silhouette advantage of these solutions (maximum $\Delta$ = 0.0006) and the high structural correspondence between the two- and three-cluster partitions confirm that the difference lies in the number of clusters rather than in how observations are assigned, further validating the robustness of the selected hyperparameters.

We also manually reviewed the 143 outlier definitions to determine whether they constituted a semantically coherent but small cluster or were genuinely outlier content. Manual inspection confirmed that the 143 definitions are semantically distinct from the three main clusters and predominantly originate from medical and clinical training contexts. The vast majority of these definitions describes operative autonomy in surgical residency training, that is, the degree to which a trainee can independently perform procedures with decreasing levels of attending supervision. Representative examples include: “a trainee’s ability to complete a procedure independently, with minimal attending supervision and participation” and “the progressive ability of trainees to perform surgical procedures and make clinical decisions with decreasing levels of supervision, ultimately leading to independent practice.” Unlike the main clusters, which theorize autonomy as a psychological capacity, a motivational orientation, or a socially situated practice, the outlier definitions do not engage with the conceptual construction of autonomy or agency but instead focus narrowly on the medical training context as a specialized operational domain. This semantic specificity places them in an isolated region of the embedding space, distant from all three cluster centroids, causing them to fall below the 5% minimum cluster-size threshold and be appropriately excluded as a substantive cluster.

To visualize the clustering results more intuitively, we projected the embeddings of all definitions into a two-dimensional space using three commonly used dimensionality-reduction methods, namely UMAP, t-SNE, and PCA, as illustrated in Supplementary Figure S2. We then compared the spatial distributions of the points under two labeling schemes: the original agency/autonomy labels and the inferred cluster labels (C1/C2/C3). For both UMAP and t-SNE, the three cluster labels produced relatively distinct boundaries, whereas the agency/autonomy labels displayed substantial overlap and no clear separation. In the PCA projection, the C1/C2/C3 embeddings also showed some degree of overlap. Nevertheless, this overlap was visibly less pronounced than that observed with the agency/autonomy labels. This outcome can be attributed to the fact that the two-dimensional PCA projection accounted for only 14.25% of the total variance and thus preserved substantially less information than the nonlinear methods. Collectively, these visualizations indicate that the clustering structure captures the semantic organization of the definitions more effectively than the original agency/autonomy distinction.

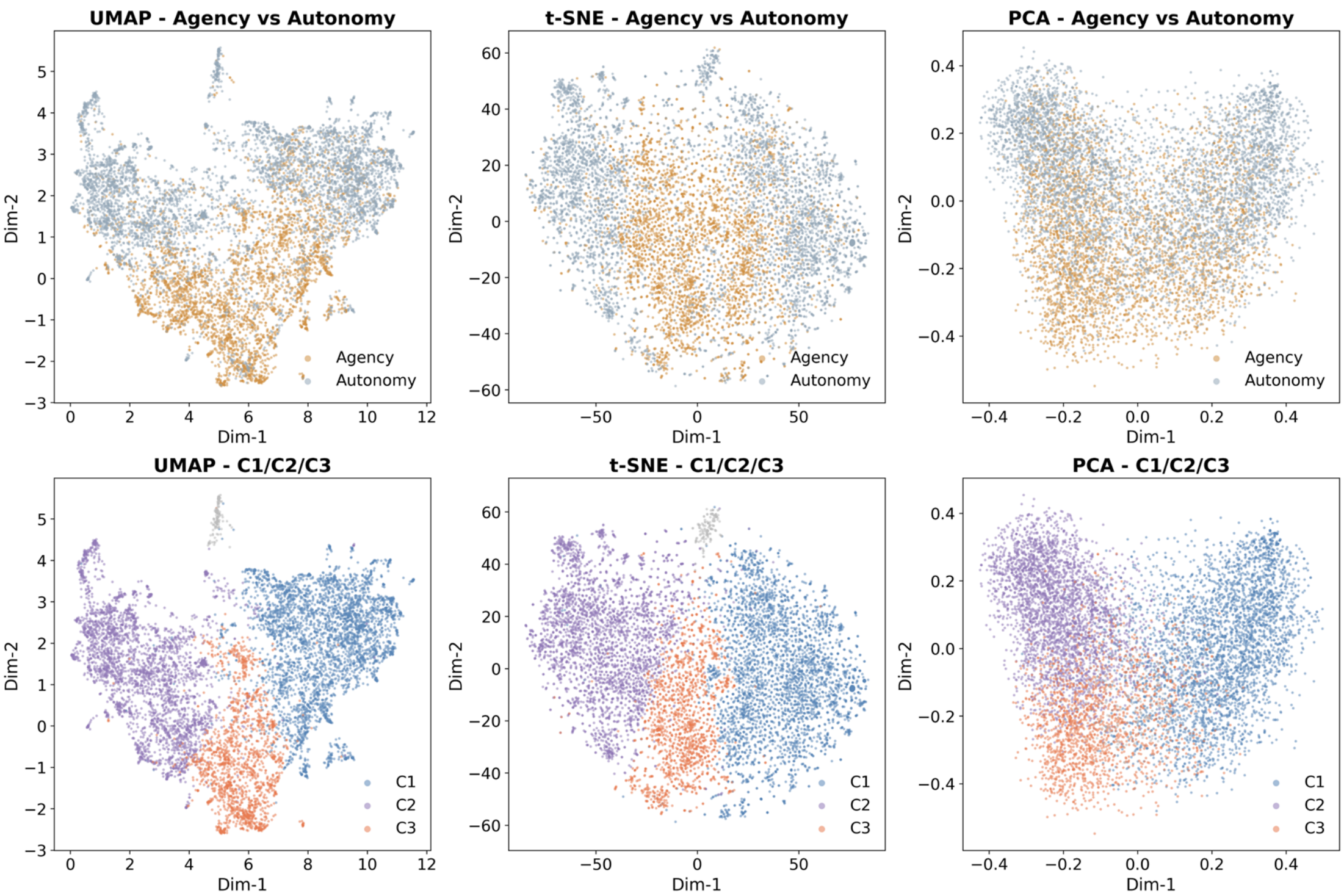

**Supplementary Figure S2.** Two-dimensional visualizations of definition embeddings using UMAP, t-SNE, and PCA. The top row shows the distributions colored by the original agency or autonomy labels, and the bottom row shows the same embeddings colored by the inferred cluster labels

C1/C2/C3. Gray points indicate outliers identified in the cluster solution.

Based on the best clustering solution, we also identified the top five definitions for each of the three clusters in terms of semantic representativeness (i.e., those with the highest cosine similarity to the cluster centroid) and the top five definitions with the highest repetition counts across the corpus, as presented in Supplementary Table S1. These definitions capture the most classic and widely adopted conceptualizations of learner agency and autonomy, including "the ability to take charge of one's own learning" (Holec, 1981), "a capacity for detachment, critical reflection, decision-making, and independent action" (Little, 1991), and "the socioculturally mediated capacity to act" (Ahearn, 2001). These representative definitions across the three clusters confirm that the clustering solution yields semantically meaningful groupings.

**Supplementary Table S1.** Most representative and most frequently cited definitions for each cluster.

| Cluster | Construct | Definition | Similarity | Repeat Times |
|---|---|---|---|---|
| Panel A: Most semantically representative definitions of each cluster | | | | |
| C1 | autonomy | the learner's ability to take control of their own learning process | 0.875 | - |
| C1 | agency | students' ability to take an active and self-directed role in their own learning process | 0.866 | - |
| C1 | autonomy | learners' ability to take charge of their own learning | 0.865 | - |
| C1 | autonomy | the learners' ability to take charge of their own learning | 0.865 | - |
| C1 | autonomy | the learners' ability to take control of their learning | 0.863 | - |
| C2 | autonomy | the ability to act of one's own volition and feel in control of one's behaviors and goals | 0.847 | - |
| C2 | autonomy | a feeling of choice and control, where one acts with their own interests and values in mind | 0.842 | - |
| C2 | autonomy | a sense of volition and self-direction | 0.836 | - |
| C2 | autonomy | the sense of volition and choice individuals experience in their actions, allowing them to align their behaviors with their values, interests, and preferences | 0.829 | - |
| C2 | autonomy | the desire for self-direction and the capacity to exercise one's volition in decision-making | 0.829 | - |
| C3 | agency | a 'temporally embedded process of social engagement', where students' ability to act is shaped by the interplay of past experiences, current contexts, and future aspirations | 0.793 | - |
| C3 | agency | a socio-culturally mediated capacity, emerging from the interplay of past identities, present constraints, and future aspirations | 0.787 | - |
| C3 | agency | an individual's "socioculturally mediated capacity to act," to design plans for change, and to create opportunities for learning | 0.784 | - |
| C3 | agency | the socio-culturally mediated capacity to act, recognizing that | 0.784 | - |
| | | it is "a relationship that is constantly co-constructed and renegotiated with those around the individual and with the society at large" | | |
| C3 | agency | a relational, situated capacity to act meaningfully within constrained structures | 0.782 | - |
| Panel B: Most frequently repeated definitions of each cluster | | | | |
| C1 | autonomy | the ability to take charge of one's own learning | 0.842 | 61 |
| C1 | autonomy | the ability to take charge of one's learning | 0.842 | 19 |
| C1 | autonomy | the ability to take charge of one's own learning | 0.841 | 17 |
| C1 | autonomy | "the ability to take charge of one's own learning" | 0.817 | 11 |
| C1 | autonomy | the ability to take charge of one's learning | 0.839 | 6 |
| C2 | autonomy | a capacity for detachment, critical reflection, decision-making, and independent action | 0.710 | 8 |
| C2 | autonomy | experiencing a sense of volition | 0.720 | 5 |
| C2 | autonomy | to feel self-determined in one's actions rather than feeling controlled | 0.750 | 4 |
| C2 | agency | the capability of individual human beings to make choices and act on these choices in a way that makes a difference in their lives | 0.742 | 4 |
| C2 | autonomy | a capacity – for detachment, critical reflection, decision-making, and independent action | 0.676 | 4 |
| C3 | agency | the socioculturally mediated capacity to act | 0.750 | 19 |
| C3 | agency | 'the socioculturally mediated capacity to act' | 0.741 | 5 |
| C3 | autonomy | the competence to develop as a self-determined, socially responsible and critically aware participant in (and beyond) educational environments, within a vision of education as (inter)personal empowerment and social transformation | 0.687 | 5 |
| C3 | agency | a student's experience of having access to or being empowered to act through personal, relational, and participatory resources, which allow him/her to engage in purposeful, intentional, and meaningful action and learning in study contexts | 0.729 | 4 |
| C3 | agency | the socio-culturally mediated capacity to act | 0.767 | 3 |

Note: "Similarity" refers to the cosine similarity between the embedding of a given definition and the cluster centroid (the mean of the embedding vectors within that cluster). "Repeat Times" indicates the number of times that particular wording appeared in the corpus.

The representative definitions exhibited a clear asymmetry across clusters. In C1 and C2, the most representative definitions were autonomy definitions, whereas in C3, the most representative definition was an agency definition. To further explore this asymmetry, we examined the distribution of cosine similarity values between the autonomy and agency constructs and the cluster centroid within each cluster. Because cluster centroids can be influenced by sample size, we constructed centroids under three conditions, namely the overall cluster centroid, the agency-only centroid, and the autonomy-only centroid. The distributions of definition-to-centroid similarities under these three conditions are presented in Supplementary Figure S3.

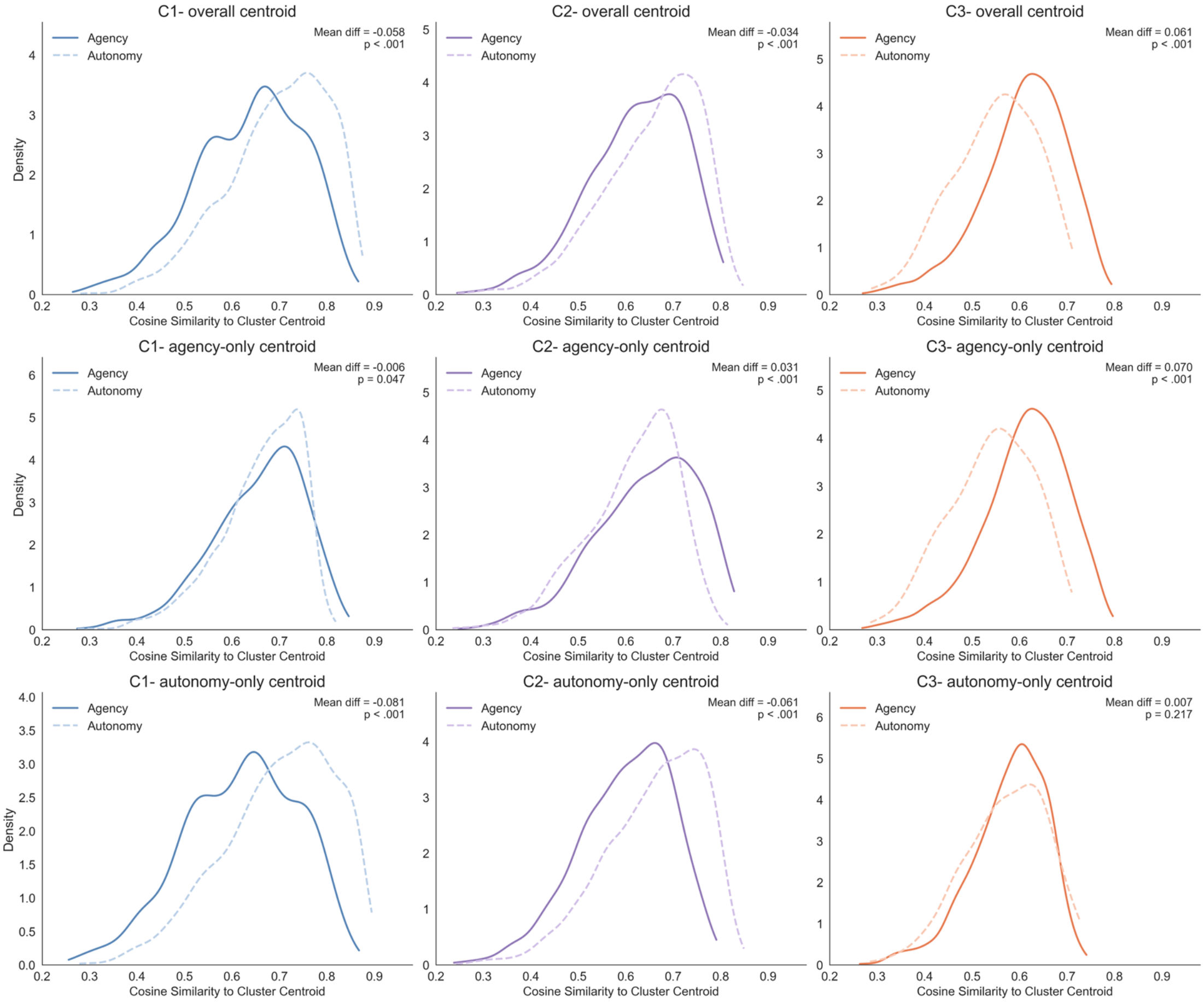

**Supplementary Figure S3.** Similarity distributions of autonomy and agency constructs relative to cluster centroids across the three clusters. For each cluster, we calculated the cosine similarity between definition embeddings and the corresponding centroid, and compared the distributions for autonomy and agency definitions under three centroids, namely the overall cluster centroid, the agency-only centroid, and the autonomy-only centroid.

Across the three clusters, when the overall cluster centroid was used, agency definitions were less similar to the centroid than autonomy definitions in C1 (agency minus autonomy = −0.058, $p < .001$) and C2 (−0.034, $p < .001$), but more similar in C3 (0.061, $p < .001$). This pattern mirrors the asymmetry described above. To examine the internal coherence of agency constructs, we turned to the agency-only centroid. In C1, even when the centroid was built exclusively from agency definitions, the agency-minus-autonomy difference remained slightly negative (−0.006, $p = .047$), meaning that agency definitions were not closer to their own centroid than autonomy definitions were. This suggests that agency definitions within C1 lack a stable, centrally organized semantic structure. To probe the organization of C3 in parallel fashion, we used the autonomy-only centroid. Under this centroid, the difference between agency and autonomy was negligible and non-significant (0.007, $p = .217$), indicating that autonomy definitions did not converge around their own centroid. In C2, the differences under the two single-construct centroids pointed in opposite directions (autonomy-only centroid: agency minus autonomy = −0.061, $p < .001$; agency-only centroid: agency minus autonomy = 0.031, $p < .001$), reflecting that both constructs preserve distinguishable semantic structures while blending into an integrated space. Together, these results show that C1 is predominantly constituted by autonomy, C3 is predominantly anchored by agency, and C2 represents a cluster in which autonomy and agency retain their individual semantic signatures yet merge into a joint conceptual organization.

## Validation of the Cluster Analysis of Definitions

To assess the robustness of the full-sample clustering solution, we repeated the same tuning and clustering procedure separately for autonomy- and agency-related definitions after deduplication by definition text. For the autonomy subset, 5,272 records were reduced to 5,015 unique definitions; for the agency subset, 3,682 records were reduced to 3,585 unique definitions. In each case, we searched over the same hyperparameter space used in the full-sample analysis and retained the best-performing solution based on silhouette scores computed in the original embedding space.

For the autonomy subset, the best-performing solution was a hierarchical partition using UMAP with n_neighbors = 30, n_components = 150, and min_dist = 0.05, followed by agglomerative hierarchical clustering with average linkage, an initial target of k = 6, and a minimum cluster ratio of 0.05 (silhouette = 0.176). Under this criterion, the retained mapped output ($n$ = 5,272) contained two substantive clusters (2,804 and 1,811 definitions) and outliers (657 definitions in total). For the agency subset, the best-performing solution was a K-means partition using UMAP with n_neighbors = 50, n_components = 100, and min_dist = 0.05, followed by K-means with k = 3 (silhouette = 0.084), which slightly outperformed the best hierarchical candidate (silhouette = 0.083). In the mapped agency corpus ($n$ = 3,682), the three clusters contained 1,084, 1,592, and 1,006 definitions.

We then compared each subset-specific optimal clustering solution with the corresponding full-sample hierarchical solution using overlapping deduplicated definitions only. As shown in Supplementary Figure S4, the autonomy-only solution remained closely aligned with the full-sample structure, while the agency-only solution showed partial but weaker one-to-one correspondence with the full-sample clusters. Under this configuration, overlap-based agreement was Normalized Mutual Information (NMI) = 0.671 and Adjusted Rand Index (ARI) = 0.757 for autonomy, and NMI = 0.604 and ARI = 0.610 for agency.

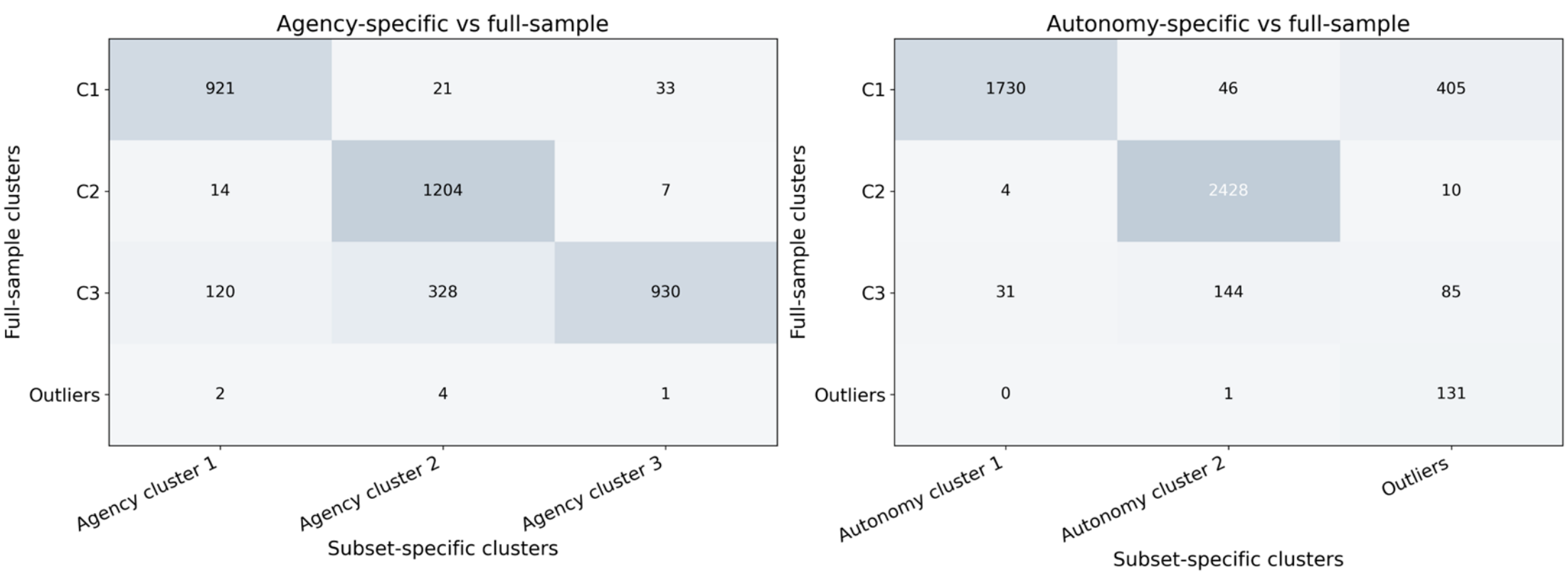

**Supplementary Figure S4.** Correspondence between the full-sample clustering solution and the autonomy-only and agency-only clustering solutions.

## Scale Extraction

After extracting definitions from full-text articles, we then conducted a dedicated scale-extraction procedure to construct a scale corpus. The measurement of agency and autonomy in educational research is methodologically heterogeneous. Some instruments were developed specifically to assess these constructs, whereas others include agency or autonomy as subscales within broader inventories, such as the autonomy subscale of the Basic Psychological Need Satisfaction and Frustration Scale (Chen et al., 2015). As a result, many relevant instruments are not readily recoverable through keyword searches for agency or autonomy alone. We therefore adopted a corpus-driven discovery strategy to identify the relevant scale-development literature as comprehensively as possible.

Starting from the 14,611 full-text articles in our literature corpus, we used the GLM-4.7 large language model to identify all articles that employed a scale to measure learner agency or autonomy and to extract the bibliographic source of each instrument used. When an article reported developing its own scale, that article itself was treated as a candidate scale-source article. When an article cited an external instrument, the cited reference was recorded as the candidate source. This automated scan identified 2,720 instrument citations across 1,920 articles. Many instruments or subscales appeared repeatedly across studies. For example, one version of the autonomy subscale of the Basic Psychological Need Satisfaction and Frustration Scale (Chen et al., 2015) was cited by 84 articles in our corpus.

We then resolved DOIs for the extracted citations via the Crossref and OpenAlex APIs and deduplicated the resulting records to obtain 939 unique candidate scale-source articles. To improve coverage of recently developed instruments that had not yet been cited within our literature corpus, we conducted supplementary manual searches on Google Scholar and related platforms, adding 12 additional references. This produced a total of 951 candidate scale-source articles. Using the same multi-channel retrieval procedure described above and manual completion, we retrieved full texts for 905 of these articles.

After retrieving the full texts of the candidate scale-source articles, we applied the GLM-4.7 model to conduct an initial screening based on three criteria: (1) the article provided complete scale items, (2) the scale or at least one of its subscales specifically measured learner or student agency or autonomy, and (3) an English version of the scale was available. A total of 383 articles met all three criteria and were retained for subsequent scale extraction.

From the retrieved full texts, we used the GLM-4.7 model to extract the final version of all scales or subscales used to measure agency or autonomy. When only a single subscale within a broader instrument targeted agency or autonomy, only that subscale was retained for analysis. For each qualifying scale or subscale, the model recorded the scale name, subscale name, item wording, reported reliability coefficient, and reverse-coding status for each item. We then excluded scales or subscales whose names did not explicitly contain strings beginning with "autonom*" or "agen*". The resulting scale corpus comprised 2,700 items drawn from 268 articles. On the basis of scale or subscale names, these items were further classified as autonomy items or agency items, yielding 1,916 autonomy items and 784 agency items.

We evaluated extraction quality through two complementary checks, an automated assessment of extraction fidelity and a manual expert review of a random sample. First, to assess extraction fidelity, we applied a text-matching procedure analogous to that used for definition extraction, in which each extracted item was automatically searched against the full text of its source article. Overall, 95.7% of extracted entries matched the source text verbatim. Manual inspection of the remaining 4.3% confirmed that all unmatched items were faithfully extracted from the original text, without substantive alteration of meaning. Manual inspection identified no entries as potential hallucinations produced by the large language model. Second, we randomly sampled 200 extracted items for independent review by one expert in the field of education against the original source texts. All 200 items were confirmed to reflect behaviors or perceptions related to learner agency or autonomy in learning contexts, consistent with the predefined extraction criteria. On this basis, we retained the full AI-extracted item set for all subsequent analyses.

## Scale Item Preprocessing

Before computing embeddings, we normalized the wording of all extracted scale items so that semantic comparisons would be less affected by superficial differences in phrasing across instruments. This step was motivated by the fact that items varied widely in grammatical form, including tense, voice, reporting perspective, and sentence structure. Although such differences do not change the construct being measured, they may distort embedding-based similarity estimates: items with similar wording but different meanings may appear close, while items with similar meanings but different wording may appear far apart. In addition, many items contained discipline-specific or setting-specific references that reflected the context of scale administration rather than the substantive meaning of agency or autonomy. If left unchanged, these contextual cues may induce spurious clustering by domain or instructional setting.

To reduce these sources of bias, we standardized all items using four rules. First, items were rewritten in simple present tense and active voice, with the learner expressed as the first-person subject. Second, contextual phrases that only specified the course, activity, lesson, or administration setting were removed when they did not contribute construct-relevant meaning, whereas process-relevant qualifiers were retained in minimal form. Third, references to specific disciplines, course titles, instructional formats, or learning environments were replaced with more general expressions. For example, terms such as "mathematics," "science laboratory," or "online learning environment" were converted into domain-neutral wording. Fourth, items marked as reverse-scored were rephrased into positive-valence forms with as little lexical modification as possible. These transformations were carried out using the GLM-4.7 large language model. The quality of preprocessing was subsequently reviewed by two domain experts in the field of education, who confirmed that the revised items preserved the original meaning while standardizing item phrasing across instruments.

One concern is that the preprocessing itself, by altering the wording of items, might shape the embedding-based results and the conclusions drawn from them rather than merely removing superficial noise. To address this, we repeated the full pipeline on the unprocessed items as a robustness check (see the *Clustering Analysis of Scale Items* section). In brief, although the resulting cluster partition was not identical to that obtained from the processed items, these differences did not affect the conclusions reported in the Results section, *Scales Reproduce Jingle–Jangle Patterns and Show Limited Content Validity*.

## Embedding Computation and Validation of Scale Items

After preprocessing, we computed embeddings for all standardized items using OpenAI's *text-embedding-3-large* model. To evaluate whether these semantic embeddings provided a valid representation of the psychometric structure of agency and autonomy instruments, following the validation approach of Wulff and Mata (2025), we examined whether statistics derived from item embeddings corresponded to established properties of the original scales. The underlying rationale was that if embeddings meaningfully captured semantic relations among items, then subscales with greater conceptual coherence should also exhibit greater coherence in embedding space, and the internal boundaries between subscales should likewise be reflected in the geometry of the embedding space.

For each subscale containing $N$ items, we first computed the mean pairwise cosine similarity among all item embeddings within that subscale:

$$S_j = \frac{1}{N_j(N_j - 1)} \Sigma_{m,n(m \neq n)} \frac{E_m \cdot E_n}{||E_m||\;||E_n||}$$

where $E_m$ and $E_n$ denote the embedding vectors of items $m$ and $n$, respectively. We then transformed this average similarity into an embedding-based estimate of Cronbach's alpha using the Spearman–Brown formula:

$$\alpha_{Embedding} = \frac{N_j S_j}{1 + (N_j - 1) S_j}$$

Reported internal consistency coefficients were extracted from the source articles using GLM-4.7, with the model instructed to return both the numerical value and the exact passage from which it was taken. To minimize the possibility of hallucinated values, we retained only those coefficients that could be directly verified against verbatim source text. Comparing the resulting reported alpha values with the embedding-based estimates showed positive correlations in both domains: $r$ = .271 ($p$ = .003, $n$ = 116) for autonomy subscales and $r$ = .355 ($p$ = .002, $n$ = 71) for agency subscales (see Supplementary Figure S5). These results indicate that semantic similarity among embedded items tracks a meaningful component of the empirical internal consistency of agency and autonomy measures.

We also evaluated whether embeddings preserved the internal dimensional organization of multi-subscale instruments. For each subscale, we compared the distribution of cosine similarities among items within the same subscale with the distribution of cosine similarities between those items and items belonging to other subscales of the same instrument. This comparison was summarized as a structural fidelity $z$-score:

$$Fidelity_j = Zscore(Sim(m,n)_{m \in \mathcal{D}_j, n \in \mathcal{D}_j, m \neq n}, Sim(m,k)_{m \in \mathcal{D}_j, k \notin \mathcal{D}_j})$$

where $Sim(\cdot,\cdot)$ denotes cosine similarity and $\mathcal{D}_j$ is the set of items belonging to subscale $j$. Larger positive values indicate that items in a subscale are more semantically similar to one another than to items in neighboring subscales, as would be expected if embeddings preserve the intended dimensional structure of the instrument. Using $z > 2$ as an indicator of clear within-subscale separation, 68.8% of autonomy subscales and 82.9% of agency subscales met this criterion (see Supplementary Figure S5). Together, the positive correspondence with reported internal consistency and the generally strong structural fidelity scores support the validity of semantic embeddings as a basis for analyzing the measurement structure of agency and autonomy in the current study.

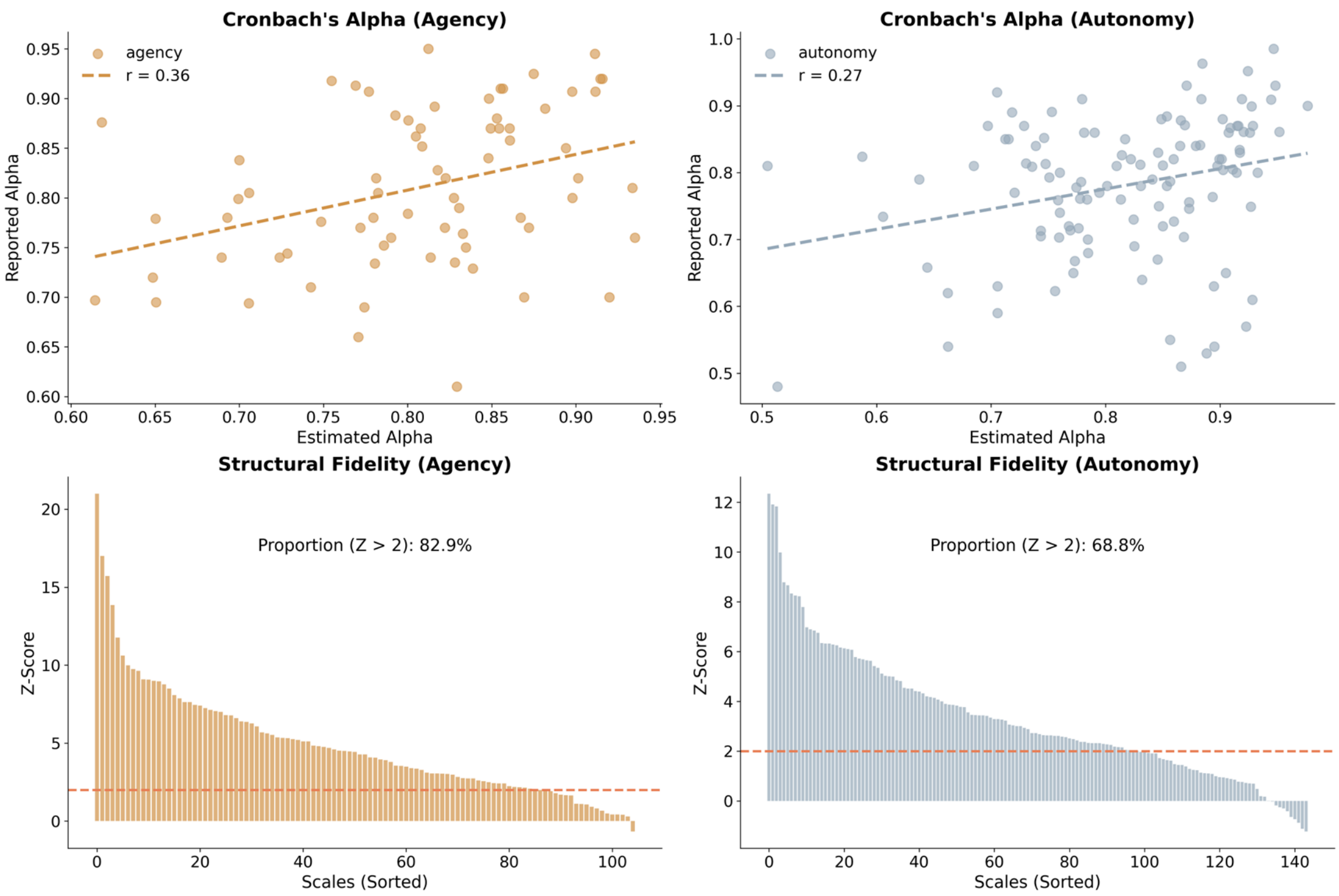

**Supplementary Figure S5.** Embedding-based validation of the psychometric structure of agency and autonomy instruments. Top panels display the correspondence between embedding-estimated and reported Cronbach's $\alpha$ values for agency and autonomy subscales. Bottom panels display the distribution of structural fidelity z-scores across agency and autonomy subscales, with dashed lines marking the threshold of $z = 2$.

## Clustering Analysis of Scale Items

We conducted clustering analysis on the embedded scale items using the same pipeline and hyperparameter search space as in the definition analysis. The full corpus contained 2,700 items. Prior to tuning, exact duplicate items were collapsed based on identical item wording, reducing the tuning set to 2,586 unique items. Hyperparameter search was then carried out over the same UMAP reduction settings and clustering models used for the definition clustering, with model selection based on silhouette score in the original embedding space. Among the hierarchical candidates, the best-performing solution used UMAP with n_neighbors = 50, n_components = 100, and min_dist = 0.05, followed by agglomerative hierarchical clustering with average linkage, k = 2, and a minimum cluster ratio of 0.03. This solution achieved the highest original-space silhouette score among the retained hierarchical item-clustering candidates (0.0996). We also evaluated K-means solutions over the same reduction grid, but these performed less well than the hierarchical solutions. The optimal retained solution was therefore a two-cluster hierarchical clustering model. The two clusters were labeled S1 and S2. In the full item corpus, S1 contained 2,183 items (autonomy = 1,528; agency = 655), whereas S2 contained 517 items (autonomy = 388; agency = 129).

Following the clustering of scale items, we further examined whether the construct labels of agency and autonomy maintained clear semantic boundaries within the scale-item embedding space. We computed pairwise cosine similarities among three pair types, including autonomy-autonomy, agency-agency, and autonomy-agency pairs, across all scale items combined and separately within each item cluster (S1 and S2). As shown in Supplementary Figure S6, the UMAP projection revealed substantial spatial overlap between agency and autonomy items, with no visually discernible separation. Quantitative analysis corroborated this observation. Across all scale items, the mean similarity of agency-agency pairs ($M$ = 0.298) was significantly lower than that of the cross-construct autonomy-agency pairs ($M$ = 0.309, $\Delta = -0.011$, Cohen's $d = -0.110$, permutation $p < .001$). This finding indicates a jingle-jangle pattern, in which items sharing the same construct label are not more semantically coherent than items spanning different labels. This pattern persisted within both clusters. In S1, agency-agency pairs again showed lower mean similarity than autonomy-agency pairs ($\Delta = -0.014$, $t = -56.52$, Cohen's $d = -0.134$, permutation $p < .001$). In S2, the same directional trend was observed ($\Delta = -0.007$, $t = -5.42$, Cohen's $d = -0.065$), though it was not significant (permutation $p = .322$). These results suggest that the jingle-jangle fallacy observed at the definitional level extends to the operationalization of these constructs in psychometric measurement.

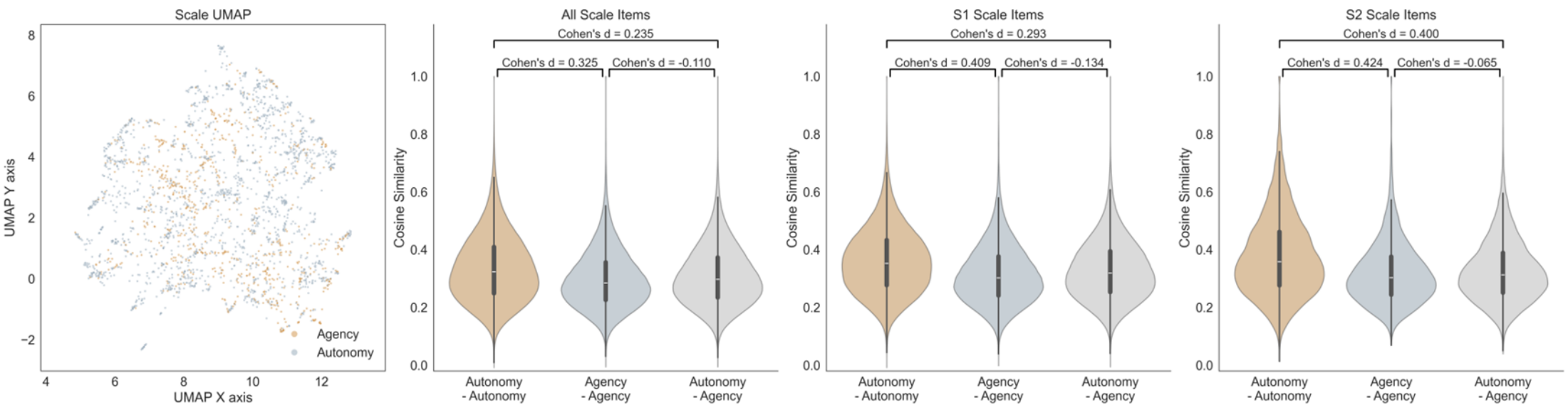

**Supplementary Figure S6.** Semantic embedding patterns of agency and autonomy scale items. The leftmost panel displays the UMAP projection of the semantic embeddings of scale items, with agency and autonomy items represented as points in distinct colors. The three right panels show violin plots depicting the distributions of cosine similarity across three pair types, autonomy-autonomy, agency-agency, and autonomy-agency, for the overall scale items (all scale items), as well as for two separate item clusters (S1 scale items and S2 scale items). Cohen's d values are annotated above each comparison to indicate effect sizes.

Because item embeddings were preprocessed in the main pipeline, we additionally ran a validation analysis on unprocessed item embeddings using the same clustering hyperparameters. The best solution was obtained with UMAP (n_neighbors = 50, n_components = 100, min_dist = 0.0) and average-linkage hierarchical clustering (n_clusters = 5, min_cluster_ratio = 0.03), yielding a silhouette of 0.080. In the full unprocessed corpus, the final assignment comprised two major clusters plus an outlier cluster (S1 contains 2,012 items; S2 contains 592 items; outliers contains 96 items). Compared with the processed solution, agreement was moderate (NMI = 0.295, ARI = 0.446), and the differences mainly arose around the S2 boundary. Nevertheless, cross-domain similarity patterns were stable. For unprocessed items, S1 remained most similar to C1 (mean cosine = 0.249, $SD$ = 0.096), and S2 remained most similar to C2 (mean cosine = 0.213, $SD$ = 0.093), while both clusters showed weaker similarity to C3 (S1-C3: $M$ = 0.158, $SD$ = 0.073; S2-C3: $M$ = 0.155, $SD$ = 0.068). These results support the conclusion that current scales remain relatively under-aligned with C3 without preprocessing.

## Operationalization of Constructs

To examine whether different definition clusters were associated with different empirical operationalization methods, we coded how learner or student agency and autonomy were operationalized in the included studies. We used GLM-4.7 to assist full-text screening and coding of the source articles corresponding to the collected definitions. The coding framework covered six broad types of operationalization commonly used in educational research. Because some studies used multiple methods to measure the same construct or adopted mixed-methods designs, multiple codes were allowed where appropriate. Studies for which the operationalization could not be reliably identified from the available text were treated as uncoded.

***Scale-based quantitative operationalization*** referred to studies that measured learner or student agency or autonomy using a validated scale, either developed within the study or adopted from prior research.

***Behavioral quantitative operationalization*** referred to studies that used quantitative analysis without an identifiable validated scale. In these studies, the focal construct was operationalized through behavioral coding, frequency counts, log data, numerical indicators, test scores, or other behavioral quantitative measures.

***Experimental or intervention-based operationalization*** referred to studies in which learner or student agency or autonomy was operationalized through experimental manipulation, quasi-experimental comparison, intervention, treatment, or pretest-posttest design.

***Interview- or focus-group-based qualitative operationalization*** referred to studies that examined learner or student agency or autonomy through interviews, focus groups, stimulated recall, or other spoken narrative data.

***Observation- or ethnography-based qualitative operationalization*** referred to studies that examined the focal construct through classroom observation, participant observation, field notes, ethnographic inquiry, video observation, or similar forms of observational tracking.

***Text-, discourse-, or document-based qualitative operationalization*** referred to studies that analyzed learner or student agency or autonomy primarily through textual, written, discourse-based, or documentary materials, such as learner diaries, reflective journals, written narratives, online posts, forum entries, curriculum texts, policy documents, discourse analysis, conversation analysis, or document analysis.

Across the 4,976 source articles in the corpus, scale-based quantitative operationalization was the most common pattern, corresponding to 1,587 articles (31.9%). Interview- or focus-group-based qualitative operationalization was the second most common, corresponding to 1,263 articles (25.4%). Text-, discourse-, or document-based qualitative operationalization accounted for 789 articles (15.9%), and observation- or ethnography-based qualitative operationalization accounted for 593 articles (11.9%). Experimental or intervention-based operationalization corresponded to 435 articles (8.7%), whereas behavioral quantitative operationalization was identified for 397 articles (8.0%). In addition, 1,312 articles (26.4%) could not be reliably assigned an operationalization code from the available text, primarily because agency or autonomy was not empirically operationalized or because the measurement procedure was not clearly described.

However, the two constructs showed different operationalization profiles. Among autonomy articles ($n$ = 3,051), scale-based quantitative operationalization was the most common ($n$ = 1,376, 45.1%), followed by interview- or focus-group-based qualitative operationalization ($n$ = 515, 16.9%), text-, discourse-, or document-based qualitative operationalization ($n$ = 293, 9.6%), experimental or intervention-based operationalization ($n$ = 273, 8.9%), behavioral quantitative operationalization ($n$ = 252, 8.3%), and observation- or ethnography-based qualitative operationalization (n = 186, 6.1%). Among agency articles ($n$ = 2,138), interview- or focus-group-based qualitative operationalization was the most common ($n$ = 812, 38.0%), followed by text-, discourse-, or document-based qualitative operationalization ($n$ = 531, 24.8%), observation- or ethnography-based qualitative operationalization ($n$ = 437, 20.4%), scale-based quantitative operationalization ($n$ = 276, 12.9%), experimental or intervention-based operationalization ($n$ = 179, 8.4%), and behavioral quantitative operationalization ($n$ = 161, 7.5%). Note that some articles addressed both constructs, so the two subsets overlap and their counts exceed the total corpus size.

Furthermore, the three definition clusters showed distinct operationalization profiles. Among C1 articles ($n$ = 2,433), scale-based quantitative operationalization was the most common ($n$ = 993, 40.8%), followed by interview- or focus-group-based qualitative operationalization ($n$ = 523, 21.5%), text-, discourse-, or document-based qualitative operationalization ($n$ = 266, 10.9%), observation- or ethnography-based qualitative operationalization ($n$ = 204, 8.4%), experimental or intervention-based operationalization ($n$ = 198, 8.1%), and behavioral quantitative operationalization ($n$ = 135, 5.5%). Among C2 articles ($n$ = 2,338), scale-based quantitative operationalization was also the most common ($n$ = 762, 32.6%), followed by interview- or focus-group-based qualitative operationalization ($n$ = 562, 24.0%), text-, discourse-, or document-based qualitative operationalization ($n$ = 408, 17.5%), observation- or ethnography-based qualitative operationalization ($n$ = 276, 11.8%), experimental or intervention-based operationalization ($n$ = 249, 10.7%), and behavioral quantitative operationalization ($n$ = 230, 9.8%). In contrast, among C3 articles ($n$ = 1,150), interview- or focus-group-based qualitative operationalization was the most common ($n$ = 474, 41.2%), followed by text-, discourse-, or document-based qualitative operationalization ($n$ = 290, 25.2%), observation- or ethnography-based qualitative operationalization ($n$ = 268, 23.3%), scale-based quantitative operationalization ($n$ = 94, 8.2%), experimental or intervention-based operationalization ($n$ = 69, 6.0%), and behavioral quantitative operationalization ($n$ = 65, 5.7%). Note that articles contributing definitions to multiple clusters were counted in each, so the cluster-level counts likewise exceed the total article count.

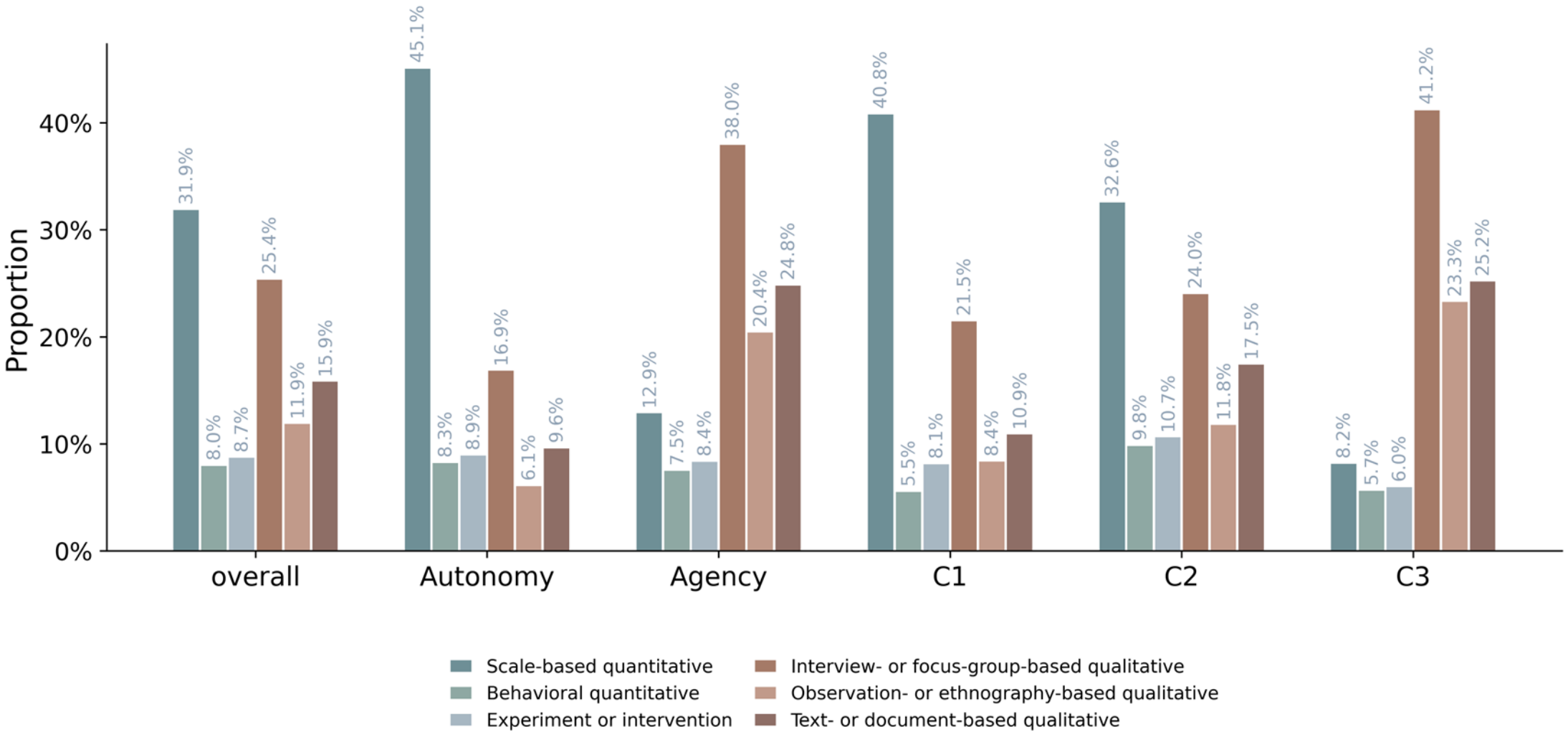

**Supplementary Figure S7.** Grouped bar chart showing the proportions of six operationalization types across the overall corpus and the three clusters. Proportions were calculated as the number of studies using a given method divided by the total number of studies in that set. Because a single

article could employ multiple operationalization methods or none were identified, the proportions may sum to more or less than 100%.

## Subsample Analysis of Generative AI Related Literature

The rapid diffusion of generative AI following the release of ChatGPT in late 2022 has led to a sharp increase in research on learner autonomy and agency in AI-mediated learning environments. To examine whether definitions used in this emerging literature exhibit a semantic profile distinct from that of the broader field, we conducted a dedicated subsample analysis within the 14,611-article corpus used in the main study. Because generative-AI-specific educational research was essentially absent before 2023, and pilot screening without a year restriction produced many false positives from unrelated pre-2023 occurrences of terms such as Claude and llama, we restricted the search to the 6,034 articles published in 2023 or later. Within these articles, we searched titles, abstracts, and author keywords using case-insensitive regular expressions for 22 generative-AI-related terms, including the conceptual descriptors "generative AI," "generative artificial intelligence," and "large language model"; the abbreviations "Gen AI," "GenAI," and "LLM"; the model names "GPT," "ChatGPT," "Claude," "Gemini," "Copilot," "Qwen," "DeepSeek," "ChatGLM," and "Llama"; and the AI-mediated functions or applications "AI-generated," "AI-assisted," "AI tutor," "chatbot," "human-AI," "AI agent," and "conversational agent." Articles were classified as generative-AI-related if at least one keyword was matched. Broader AI-related terms such as intelligent tutoring system and deep learning were deliberately excluded. Matching was performed on title, abstract, and keyword fields rather than full text in order to prioritize studies substantively focused on generative AI rather than those mentioning it only incidentally. This procedure identified 372 generative-AI-related articles among the 6,034 publications from 2023 onward, representing 6.2% of that subset. Of these, 140 articles contributed definitional units represented in the clustered corpus, yielding 223 definitions in total, including 113 autonomy definitions and 110 agency definitions. To construct a temporally clean comparison baseline that excludes any latent generative-AI influence, we restricted the non-GenAI comparison set to articles published before 2023, which contributed 5,303 definitions. All analyses were conducted at the definition level, and $\chi^2$ tests of cluster distribution excluded the small number of outlier-cluster definitions, leaving 220 GenAI definitions and 5,237 pre-2023 non-GenAI definitions assigned to C1 through C3.

To assess whether the findings reported in the main text (see Results section *Generative AI Research Amplifies C1 Concentration*) could be confounded by publication year, we repeated the analysis using only the non-GenAI articles published in 2023 or later as the comparison group, thereby holding the publication window constant across the two sets. This contemporaneous baseline contributed 3,428 definitions, of which 3,354 were assigned to C1 through C3 after removing outliers. Under this more stringent comparison, the overall cluster shift remained statistically significant and of comparable magnitude to the pre-2023 baseline ($\chi^2(2) = 21.09$, $p < .001$, Cramér's $V = 0.077$), with 44.6%, 36.4%, and 19.0% of contemporaneous non-GenAI definitions in C1, C2, and C3, against 60.4%, 25.0%, and 14.6% for GenAI definitions. The same held for both the autonomy-only comparison ($\chi^2(2) = 8.59$, $p = .014$, Cramér's $V = 0.066$) and the agency-only comparison ($\chi^2(2) = 15.85$, $p < .001$, Cramér's $V = 0.099$). By contrast, the difference in construct choice was not significant ($\chi^2(1) = 2.13$, $p = .145$, Cramér's $V = 0.024$). Additionally, we directly compared the pre-2023 and 2023-and-later non-GenAI definitions on their cluster distributions and found no evidence of a temporal trend: the proportions were nearly identical across periods (C1: 42.8% vs. 44.6%; C2: 37.7% vs. 36.4%; C3: 19.5% vs. 19.0%; $\chi^2(2) = 2.57$, $p = .276$, Cramér's $V = 0.017$). Together, these results indicate that the semantic shift documented here is attributable to the generative-AI research context itself rather than to a temporal change in how learner autonomy and agency are conceptualized.

A further concern is that the C1/C2/C3 partition was derived from the full corpus, which itself contains the 223 generative-AI-related definitions. If these definitions exerted disproportionate influence on cluster formation, the apparent semantic shift could be partly an artifact of using a partition that was already shaped by generative-AI content. To rule out this possibility, we re-ran the clustering pipeline on the pre-2023 subsample alone, which contains no generative-AI-related definitions. The same algorithm and hyperparameter search procedure used in the main analysis were applied, except that the upper end of the search range for UMAP's n_neighbors parameter was extended (from {10, 30, 50} to {10, 30, 50, 100, 200, 300}) to accommodate the smaller sample size, since larger neighborhoods recover the underlying manifold more stably when fewer points are available. The optimal solution on the pre-2023 subsample was again a three-cluster partition, mirroring the structure obtained from the full corpus. Comparing the two solutions over their shared definitions yielded NMI = 0.715 and ARI = 0.801, both indicating substantial agreement. This confirms that the C1 through C3 partition is temporally stable and not an artifact of generative-AI content in the corpus, and accordingly that interpreting the GenAI-related semantic shift against this partition is well-founded.

## References for Supplementary Information